\documentclass[journal]{IEEEtran}
\usepackage{graphicx}
\usepackage{amsmath,amssymb} 
\usepackage{color}
\usepackage{multirow}
\usepackage{multicol}
\usepackage{subfigure}

\definecolor{CQCColor}{rgb}{1.0,0.0,0.0}

\definecolor{CQColor}{rgb}{0,0,1.0}

\definecolor{ZYColor}{rgb}{1.0,0,1.0}

\ifCLASSINFOpdf
\else
\fi

\begin{document}
	%
	\title{SADRNet: Self-Aligned Dual Face Regression Networks for Robust 3D Dense Face Alignment and Reconstruction}
	%
	%
	%
	
	\author{Zeyu Ruan,
		Changqing Zou,~\IEEEmembership{Member,~IEEE},
		Longhai Wu,
		Gangshan Wu,~\IEEEmembership{Member,~IEEE},
		Limin Wang,~\IEEEmembership{Member,~IEEE}
		\thanks{Z. Ruan, G. Wu, L. Wang are with  the  State  Key  Laboratory for Novel Software Technology, Nanjing University, Nanjing, 210023, China (e-mail: mg1833060@smail.nju.edu.cn, gswu@nju.edu.cn, lmwang@nju.edu.cn).}
		\thanks{C. Zou is with the School of Data and Computer Science, Sun Yat-sen University, Guangzhou, 510006, China (e-mail: aaronzou1125@gmail.com).}
		\thanks{L. Wu is with the Samsung Electronics (China) R\&D Centre, Nanjing, 210012, China (e-mail: longhai.wu@samsung.com).}
	}
	
	%
	%

	\markboth{IEEE Transactions on Image Processing}%
	{Shell \MakeLowercase{\textit{et al.}}: Bare Demo of IEEEtran.cls for IEEE Journals}
	%



	\maketitle
	
	\begin{figure*}[ht]
		\begin{center}
		\includegraphics[width=1\linewidth]{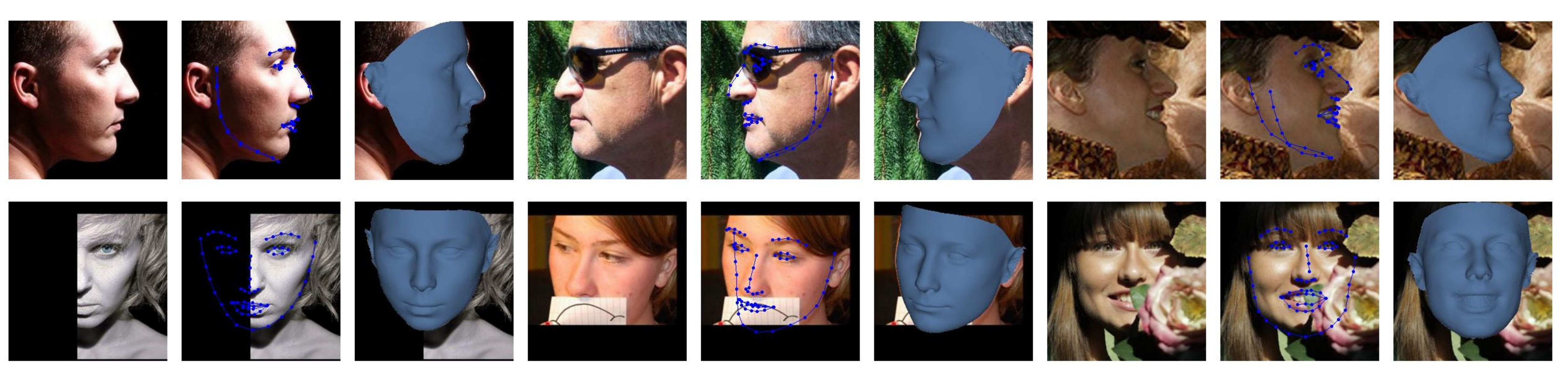}
		\end{center}
		\caption{ Illustration of the challenges of large poses (the 1st row) and occlusions (the 2nd row). The results of both face alignment and reconstruction are demonstrated. Only 68 landmarks are plotted for better view.
		}
		\label{fig:example}
	\end{figure*}
	
	\begin{abstract}
		Three-dimensional face dense alignment and reconstruction in the wild is a challenging problem as partial facial information is commonly missing in occluded and large pose face images. Large head pose variations also increase the solution space and make the modeling more difficult. Our key idea is to model occlusion and pose to decompose this challenging task into several relatively more manageable subtasks. To this end, we propose an end-to-end framework, termed as Self-aligned Dual face Regression Network (SADRNet), which predicts a pose-dependent face, a pose-independent face. They are combined by an occlusion-aware self-alignment to generate the final 3D face. Extensive experiments on two popular benchmarks, AFLW2000-3D and Florence, demonstrate that the proposed method achieves significant superior performance over existing state-of-the-art methods.
		
	\end{abstract}

	\IEEEpeerreviewmaketitle
	
	\section{Introduction}
	
	Monocular 3D face reconstruction recovers 3D facial geometry from a single-view image. Dense face alignment (e.g.,~\cite{3DDFA,PRN}) locates all facial vertices of a face model.
	They are closely related to each other, and both play important roles in broad applications such as face recognition~\cite{frjp,face_recog}, normalization~\cite{Zhu_2015_CVPR}, tracking~\cite{Xiong_2015_CVPR}, swapping~\cite{swap}, and expression recognition~\cite{express} in computer vision and graphics.
	
	The existing methods that simultaneously address 3D dense face alignment and face reconstruction (3D-DFAFR, for short) can be roughly grouped into two categories: model-based category and model-free category.
	Model-based methods infer the parameters of a parametric model
	~\cite{FLAME,Loper:2015:SSM:2816795.2818013}, such as a 3D morphable model (3DMM~\cite{Blanz:1999:MMS:311535.311556}), by solving a nonlinear optimization problem or directly regressing with convolutional neural networks (CNNs)~\cite{MMFace2019_CVPR,extreme_2018_CVPR,3DDFA,dual_attention,graphcvpr20,3ddfav2}.
	Recent work~\cite{Nonlinear_3DMM,FPS,on_learning,Tran_2019_CVPR,FPS}
	achieves high-fidelity face shape and dense face alignment by using
	nonlinear 3DMM decoders to improve its representation power.
	Rather than using a parametric model, model-free methods obtain unrestricted 3D face structure and alignment information by directly inferring the 3D position of face vertices represented in specific forms (e.g., UV map~\cite{PRN}, volume~\cite{VRN}).
	
	Although significant improvements have been achieved on the problem of 3D-DFAFR in a controlled setting during the past few years, 3D-DFAFR under unconstrained conditions is still yet to be well addressed.
	Specifically, under unconstrained conditions, as shown in Fig.~\ref{fig:example}, self occlusions caused by large pose orientation and inter-object occlusion of hair and glasses could significantly reduce the useful information in an image, making it challenging to generate good results.
	Meanwhile, large pose diversity will make it hard to distinguish the shape variations and increase the modeling difficulty.
	Previous works mainly tackle these problems by increasing the data size and diversity of training data~\cite{3DDFA,PRN,FPS,cycle_gan_2017_ICCV} or performing strong regularization on the shape~\cite{semi_2019_ICCV}.

	To addresses the challenges of 3D-DFAFR in the wild, this paper proposes an effective solution based on the following three motivations:
	(1) an occluded region in the image does not contain any face information but may negatively affect the network prediction.
	(2) the occluded part of a face can only be inferred through the global facial structure or prior knowledge.
	(3) disentangling the face pose and face shape will significantly reduce the complexity of the problem of 3D-DFAFR and thus make it more tractable.
	Based on the above motivations,  we believe {\em occlusion} and {\em pose} are two critical factors in achieving a robust 3D-DFAFR. Unfortunately, there are few 3D-DFAFR works explicitly handle with the face occlusions. The face pose estimation is also rarely discussed deeply in existing 3D-DFAFR works. Therefore, in this work, we propose to explicitly model these two factors to build a robust method for 3D-DFAFR.

	We propose a self-aligned dual face regression network (SADRNet) for robust 3D dense face alignment and face reconstruction based on the above analysis. In particular, we present a dual face regression framework to decouple face pose estimation and face shape prediction in a low computational cost manner. These two regression networks share the same encoder-decoder backbone. They are equipped with its task-specific head design for predicting pose-dependent face and pose-independent face, respectively. The pose-independent shape regression could relieve the difficulty of directly predicting the original complex face shape under various poses and thus improve the shape prediction accuracy. Besides, to tackle the occlusion issue, we devise a supervised attention mechanism to enhance the discriminative features in visible areas while suppressing the occluded region's influence. The attention mechanism could be plugged into our dual face regression framework to improve prediction accuracy. Finally, we propose an occlusion-aware self-alignment module to combine the pose-dependent and pose-independent faces to yield the final face reconstruction. In this alignment module, we only use the visible and sparse face landmarks to estimate the pose parameters, which could further improve the robustness of our SADRNet. Our solution significantly improves the robustness toward face occlusions in the wild. It achieves a considerable margin on both face reconstruction and dense face alignment. In summary, the main contributions of this paper are:
	\begin{itemize}
		\item We propose a self-aligned dual faces regression framework, which is robust to face pose variation and occlusion, for the problem of 3D-DFAFR.
		\item We propose an attention-aware mechanism for visible face region regression, which can improve the regression
		accuracy and robustness under various situations of face occlusion.
		\item The proposed end-to-end architecture is efficient and it can run at $224\times224/70$ FPS on a single GTX 1080 Ti GPU.
		\item The proposed method achieves a considerable margin on the challenging AFLW2000-3D~\cite{3DDFA} dataset and Florence 3D Faces~\cite{florence} over the state-of-the-art methods.
	\end{itemize}

	\section{Related Work}
	\subsection{3D Face Reconstruction}
	Blanz and Vetter~\cite{Blanz:1999:MMS:311535.311556} proposed 3DMM to represent a face model by a linear combination of orthogonal bases obtained by PCA. In this way, 3D face reconstruction can be formulated as 3DMM parameters regression problems. Later, Paysan et al.~\cite{BFM2009} extended the model by adding more scans and decomposing the expression bases from shape bases. A lot of earlier methods regressed the 3DMM parameters by solving a nonlinear optimization function~\cite{Lee2012,Zhu_2015_CVPR,BFM2009,fit3DMMfeat2015,estimate2005}. Thanks to the development of deep learning, some methods~\cite{Jourabloo_2016_CVPR,3DDFA,Sela_2017_ICCV,very_deep_2017_CVPR,MMFace2019_CVPR,3ddfav2,photoreal,avartame} started to use deep convolutional neural network (DCNN) architectures to learn 3DMM parameters and largely replaced traditional optimization-based methods with more accurate results and shorter running time. The self-supervised training of DCNN-based methods was implemented by exploiting a differentiable renderer~\cite{differential_2018_CVPR,ringnet,GANFIT,multiview_2019_CVPR,multiframe_2019_CVPR,accurateweak,MGC}, which alleviated the lack of 3D-supervised data and improved the generalization of networks.
	However, these model-based methods' reconstruction geometry is constrained by the linear bases with limited representation power.
	
	Some works proposed to break the limitation by using nonlinear models~\cite{disentage_2018_CVPR,Bagautdinov_2018_CVPR,Nonlinear_3DMM,Tran_2019_CVPR,dis_2019_CVPR,disentage_2018_CVPR,ref1}.
	In~\cite{on_learning}, a DCNN was used as a nonlinear 3DMM decoder. Ranjan et al.~\cite{mesh_ae_2018_ECCV} used spectral graph convolutions to learn 3D faces. Zhou et al.~\cite{FPS} presented a nonlinear 3DMM using colored mesh decoding. Guo et al.~\cite{facefromX} learn a more powerful nonlinear 3DMM  from different data sources: scanned 3D face, RGB-D images, and RGB images.
	
	Some other works directly obtained the full 3D geometry to avoid the restriction of parametric models and difficulty in pose estimation~\cite{dualcouple,PRN,mesh_ae_2018_ECCV}. Jackson et al.~\cite{VRN} proposed to use a volumetric representation of 3D face shape instead of the previously used point cloud or mesh and directly regressed the voxels. Feng et al.~\cite{PRN} mapped the mesh of a face geometry into UV position maps and then trained a light-weighted network that obtains the 3D facial geometry along with its correspondence information.
	
	Some works also combine the 3DMM-based regression and direct 3D geometry regression to improve the reconstruction performance. Chen et al.~\cite{UVdisplacement} used a 3DMM-based coarse model and a displacement map in UV space to represent a 3D face, and utilize the input image as supervision to effectively learn the facial details. Huber et al.~\cite{MMFace2019_CVPR} proposed to  estimate an intermediate volumetric geometry and finetune it with 3DMM parameters' regression.

	\subsection{Face Alignment}
	In the beginning, face alignment works aimed to locate a set of 2D facial landmarks in the image plane. Traditional works were mainly based on Active Appearance Models (AMM) and Active Shape Models (ASM)~\cite{AAM,ASM,align1,align2,align3} and considered face alignment as a model parameter optimization problem. Then cascaded regression methods~\cite{CPR1,CPR2} became popular. They iteratively refined the predictions and reached higher accuracy. With the development of deep learning, CNNs were wildly used to regress the landmarks' positions directly or predict heat maps and largely improved the performance~\cite{wingloss,cnnalign1,PengECCV16,JVCR}. Since the 2D face alignment methods have limitations on detecting invisible landmarks, the 3D face alignment problem has been widely researched in recent years. There are two major strategies for 3D face alignment: (1) separately detecting 2D landmarks and their depth~\cite{how_far,BMVC2016_86,Bulat2016TwoStageCP}, and (2) fitting a certain 3D face model~\cite{hough_ECCV16,without_cor_ECCV16} to obtain the full 3D structure to guide the 3D landmark localization.
	
	Since the methods mentioned above can handle only a limited quantity of landmarks, which is far from enough in some applications, 3D dense face alignment~\cite{defa_2017_ICCV} started to be researched. It requires methods to offer pixel-wise facial region correspondence between two face images. As the prediction target changes from a sparse set of facial landmarks to a dense set of tens of thousands of points, it is natural to solve this problem by fitting a registered 3D face model.
	
	\begin{figure*}[htbp]
		\begin{center}
			
			\includegraphics[width=1\linewidth]{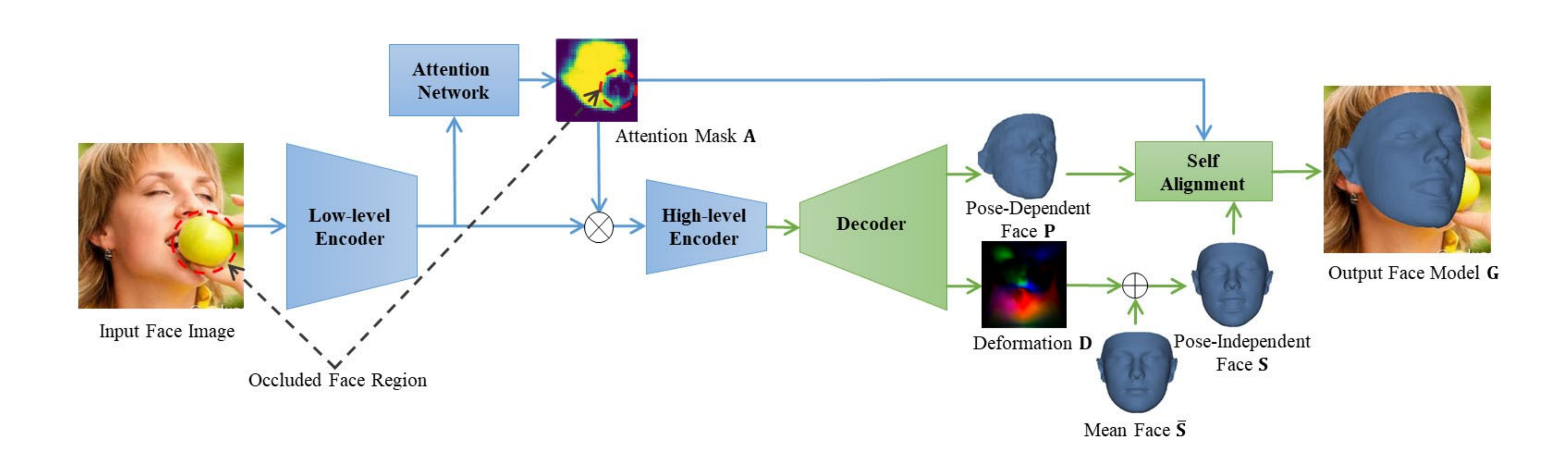}
		\end{center}
		\caption{The framework of our proposed self-aligned dual face regression network (SADRNet). $\mathbf{A}$ is the attention mask. $\mathbf{P}$ is the pose-dependent face. $\mathbf{D}$ is the face shape deformation (visualized in UV space). $\mathbf{\bar{S}}$ is the mean face template. $\mathbf{S}$ is the pose-independent face. $\mathbf{G}$ is the output face model.
		}
		\label{fig:framework}
	\end{figure*}
	
	\subsection{3D Dense Face Alignment and Face Reconstruction}
	Most model-based reconstruction approaches can be applied to dense face alignment if the face model is well registered and provides a dense correspondence~\cite{defa_2017_ICCV,Nonlinear_3DMM,MMFace2019_CVPR}. Explicit facial pose estimation is needed in these works.
	Zhu et al. proposed 3DDFA~\cite{3DDFA}, which is a representative work in this task. They used a cascaded CNN framework to regress the 3DMM parameters, including the pose parameters. The rotation angles are represented by a 4D unit quaternion for less difficulty in learning. However, their output faces are sensitive to the fluctuation of every parameter and hard to reach high precision. In ~\cite{MMFace2019_CVPR}, an ICP post-processing is incorporated to refine the regressed pose parameters, but it takes enormous extra computation. Zhou et al. learned a nonlinear 3DMM by directly using graph convolutions on face meshes and reached an extremely fast decoding speed. Nevertheless, the face pose is still obtained by direct regression like 3DDFA.
	
	Some model-free methods that directly predict 3D coordinates of facial points are also applicable to the task of dense face alignment with their face representation registered with fixed semantic meaning~\cite{PRN,VRN}. However, it is difficult for these model-free methods to handle severe occlusions and large poses since no prior knowledge or constraint is provided.
	
	Our method's final output face geometry is represented by a UV position map, which possesses a dense correspondence between the face shape and the input image.
	Unlike the methods mentioned above, we explicitly deal with object occlusions by leveraging an attention mechanism on network features. We predict a pose-independent face model to avoid the large variance in face shape brought by large poses. Instead of directly regressing the pose parameters, we perform a visibility-aware self-alignment between a pose-dependent face model and a pose-independent face model to estimate the pose of a nonlinear 3DMM. The alignment process is stable as the error caused by a single outlier is apportioned by all of the landmarks. In contrast, the error of every pose parameter is accumulated in parameter-regressing methods. Besides, the alignment process is based on two regressed faces with similar shapes, rather than the matching of landmarks and a fixed face template. Therefore, the change of face shape is taken into account in pose estimation. In this way, our method is more robust and accurate than previous works.
	
	\section{Method}
	In this section, we detail the SADRNet. We first introduce the dual face representation used in this work and the network architecture overview. We then present the occlusion-aware attention mechanism and the face fusion module. After that, we introduce the loss functions and implementation details.
	\subsection{Dual Face Regression Framework}
	\label{sec:framework}

	{\bf Facial geometry representation.}
	We assume the projection from the 3D face geometry to the 2D image is a weak perspective projection:
	
	\begin{equation}
		\mathbf{V}=\mathbf{Pr}*\mathbf{G}
	\end{equation}
	where $\mathbf{V}$ is the projected geometry on the 2D plane, $\mathbf{Pr}=\begin{bmatrix}~1&~0&~0\\~0&~1&~0\end{bmatrix}$ is the projection matrix, $\mathbf{G}\in\mathbb{R}^{3\times n}$ is the 3D mesh of a specific face with $n$ vertices.
	
	We separate the 3D face geometry into pose, mean shape, and deformation as:
	\begin{equation}
		\label{eq_1}
		\mathbf{G}=f*\mathbf{R}*\mathbf{S}+\mathbf{t},
	\end{equation}
	\begin{equation}
		\label{eq_2}
		\mathbf{S}=(\mathbf{\bar{S}}+\mathbf{D}),
	\end{equation}
	$\mathbf{S}\in\mathbb{R}^{3\times n}$ represents the pose-independent (i.e., pose-normalized) face shape,
	$\mathbf{\bar{S}}\in\mathbb{R}^{3\times n}$ is the mean shape template provided by~\cite{BFM2009} and $\mathbf{D}\in\mathbb{R}^{3\times n}$ is the deformation between $\mathbf{S}$ and $\mathbf{\bar{S}}$.
	The pose parameters consist of the scale factor $f$, the 3D rotation matrix $\mathbf{R}\in\mathbb{R}^{3\times 3}$ and the 3D translation $\mathbf{t}\in\mathbb{R}^{3}$.

	{\bf Dual face regression and analysis.}
	We propose to jointly regress two face models: a pose-independent face (the face shape) $\mathbf{S}$ and a pose-dependent face $\mathbf{P}$. And then, we use a self-alignment post-process $\phi$ to estimate face pose from $\mathbf{P}$, $\mathbf{S}$ and facial vertices' visibility information $\mathbf{Vis}$:
	\begin{equation}
		\phi(\mathbf{P},\mathbf{S},\mathbf{Vis})=f,\mathbf{R},\mathbf{t}.
	\end{equation}
	The visibility information $\mathbf{Vis}$ is explained in Sec.\ref{subsec:attention}. Based on the estimated face pose $\phi(\mathbf{P},\mathbf{S},\mathbf{Vis})$ and pose-independent face $\mathbf{S}$, we could reconstruct our final face shape $G$ via transformation defined in Eq.\ref{eq_1}.
	
	In fact, $\mathbf{G}$ and $\mathbf{P}$ have the same physical meaning, which means the ground truth of them are the same:
	\begin{equation}
		\hat{\mathbf{P}}=\hat{\mathbf{G}}.
	\end{equation}
	The difference is that $\mathbf{P}$ is obtained by direct network regression, while $\mathbf{G}$ is obtained by applying Eq.\ref{eq_1}. To distinguish them, we term $\mathbf{P}$ as pose-dependent face and $\mathbf{G}$ as face geometry.
	The learning of pose-dependent face is easy to over-fit to pose and under-fit to shape as the orientation variations bring much greater point-to-point distances than the shape variations, resulting in some implausible face shape under large pose cases. By contrast, the pose-independent face $\mathbf{S}$ does not change with the pose. Thus the network would focus on the shape characteristics and learn more details.  $\mathbf{S}$ is disentangled into the mean face shape template $\mathbf{\bar{S}}$ and the deformation $\mathbf{D}$ between the actual shape and the mean shape. Only the zero-centered $\mathbf{D}$ needs to be predicted. It further reduces the fitting difficulty. The mean face shape also serves as prior knowledge to keep the invisible facial parts plausible. By combining the shape of $\mathbf{S}$ with the pose $\phi(\mathbf{P},\mathbf{S},\mathbf{Vis})$ estimated from $\mathbf{P}$ and $\mathbf{S}$, we are able to get a much better $\mathbf{G}$, as demonstrated in experiments.
	
	{\bf UV map representation.}
	The face geometry $\mathbf{G}$, pose-independent face $\mathbf{S}$, mean face $\mathbf{\bar{S}}$, deformation $\mathbf{D}$, and pose-dependent face $\mathbf{P}$
	are transformed into UV space~\cite{PRN} as UV maps.
	UV map $\mathbf{U}$ is a 2D representation of 3D vertices on the face mesh model.
	It can be expressed as
	\begin{equation}
		\mathbf{U}(u_i,v_i)=(x_i,y_i,z_i),
	\end{equation}
	where $(x_i,y_i,z_i)$ is the 3D coordinate of vertex $i$ on the 3D face mesh and $(u_i,v_i)$ is the corresponding 2D UV coordinate.
	\begin{figure}[tbp]
		
		\centering
		\subfigure[]{
			\includegraphics[width=0.35\linewidth]{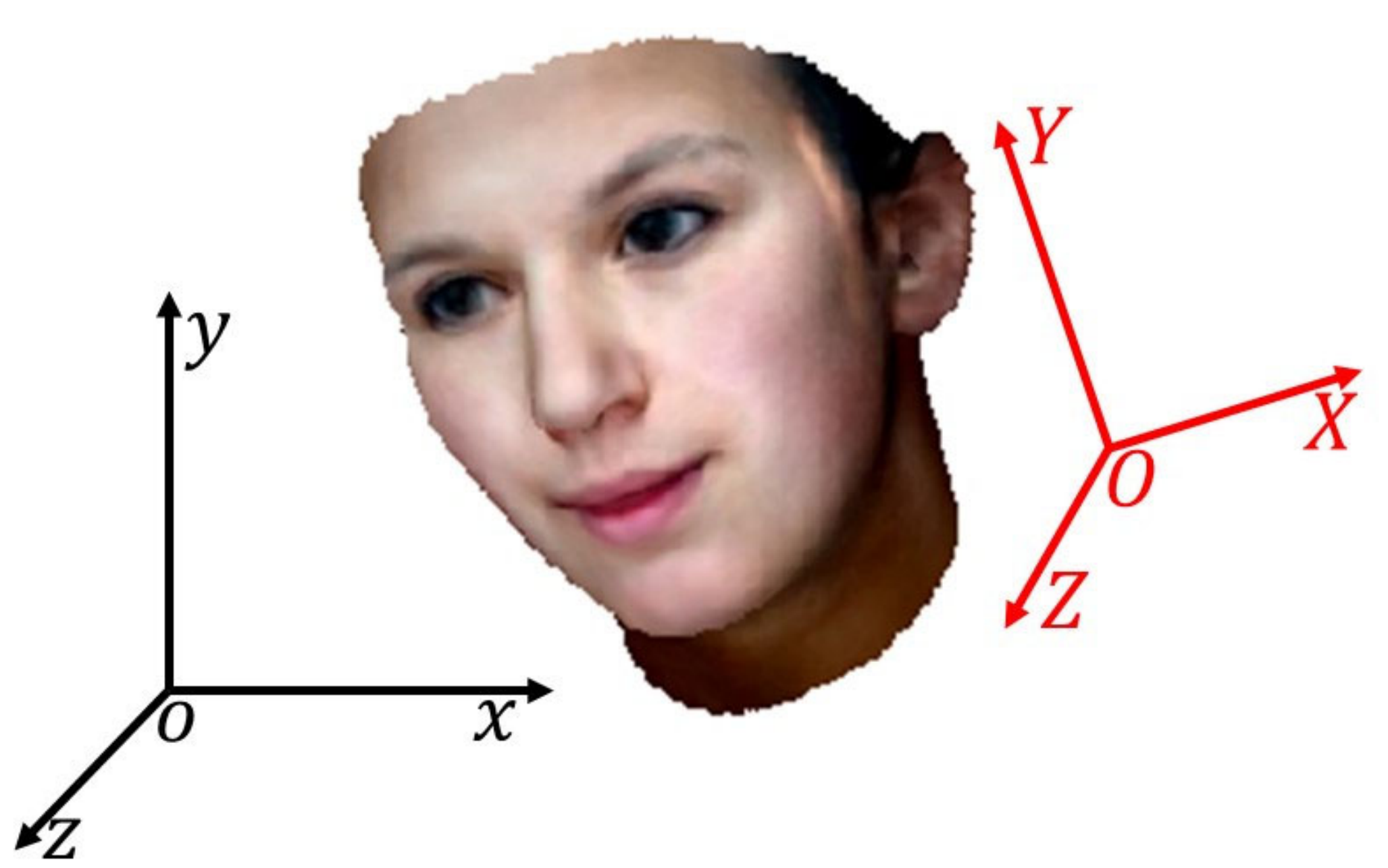}
		}
		\hspace{0mm}
		\subfigure[]{
			\includegraphics[width=0.25\linewidth]{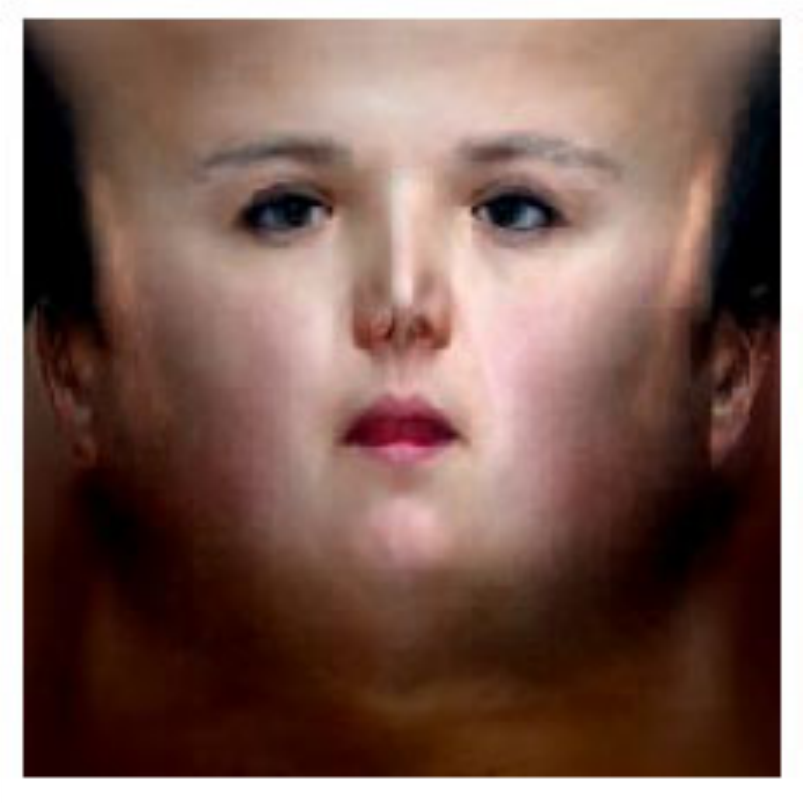}
		}
		\hspace{-2mm}
		\subfigure[]{
			\includegraphics[width=0.25\linewidth]{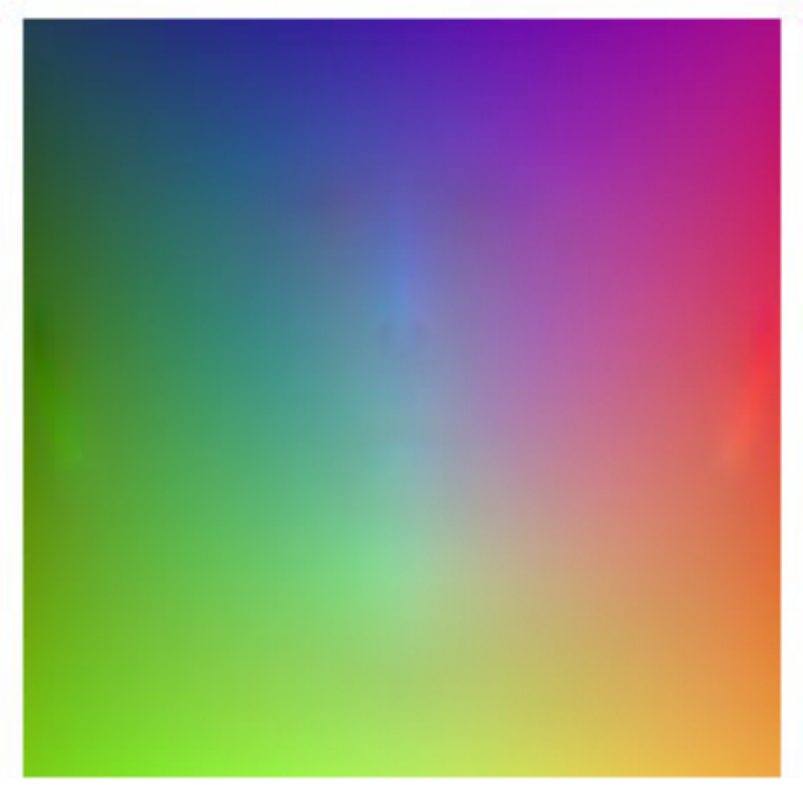}
		}
		\hspace{-2mm}
		
		\caption{Illustration the UV map of a 3D face model. Given a textured face model (a), we can compute the corresponding unwrapped texture in UV space (b), and the corresponding UV position map (c). }
		\label{fig:new_uv}
	\end{figure}
	
	The mapping relationship between the 3D object coordinate $(X_i,Y_i,Z_i)$ and the UV coordinate $(u_i,v_i)$ of a facial vertex $i$ can be formulated as
	\begin{equation}
		u_i\rightarrow \alpha_1\cdot Y_i +\beta_1,
	\end{equation}
	\begin{equation}
		v_i\rightarrow \alpha_2\cdot \arctan(\frac{X_i}{Z_i})+\beta_2,
	\end{equation}
	where $\alpha_1$, $\alpha_2$, $\beta_2$, $\beta_1$ are scaling and translation constants. The mapping relationship is computed on the mean face mesh from the Basel Face Model (BFM)~\cite{BFM2009} and applied to all the face meshes. In this way, the points in the UV map are registered to the face mesh model. Fig.~\ref{fig:new_uv} illustrates the mapping for a better understanding.
	This representation guarantees the spatial consistency between face model and UV map, i.e., spatially neighboring points on the face model are neighboring in the UV map. Since 2D UV maps can be processed by sophisticated CNNs, this ensures a great potential of the representation in applications of unconstrained situations.
	In the remainder of this paper, the face geometry, pose-independent face, mean face, deformation, and pose-dependent face are represented as UV maps.
	For clarity, the same notation is used for a thing in the two spaces (e.g., $\mathbf{G}$ is used to
	denote both the face geometry and the UV map of it).

	{\bf Network architecture.} Our self-aligned dual face regression network is an encoder-decoder-based architecture that regresses the deformation $\mathbf{D}$ and infers the pose parameters $f$, $\mathbf{R}$ and $\mathbf{t}$ to reconstruct the 3D face geometry from a single 2D face image.
	It consists of three sub-networks: encoder, attention side branch, and decoder.
	
	The encoder network contains a sequence of residual blocks that first extract low-level features, which are then fed into the attention side branch. The attention network generates an attention mask $\mathbf{A}$ with the visible face region highlighted. Then encoder network further encodes the attended features into high-level features.
	The decoder then decodes them into the pose-dependent face $\mathbf{P}$ and the deformation $\mathbf{D}$. The face shape $\mathbf{S}$ is obtained by Eq. \ref{eq_2}. The pose parameters $f$, $\mathbf{R}$ and $\mathbf{t}$ are then obtained from $\mathbf{P}$, $\mathbf{S}$ and $\mathbf{A}$ in the self-alignment module. The final face geometry $\mathbf{G}$ is generated by applying Eq. \ref{eq_1}.

	\subsection{Occlusion-Aware Attention Mechanism}
	\label{subsec:attention}
	\begin{figure}[tbp]
		
		\centering
		\subfigure[]{
			\includegraphics[width=0.155\linewidth]{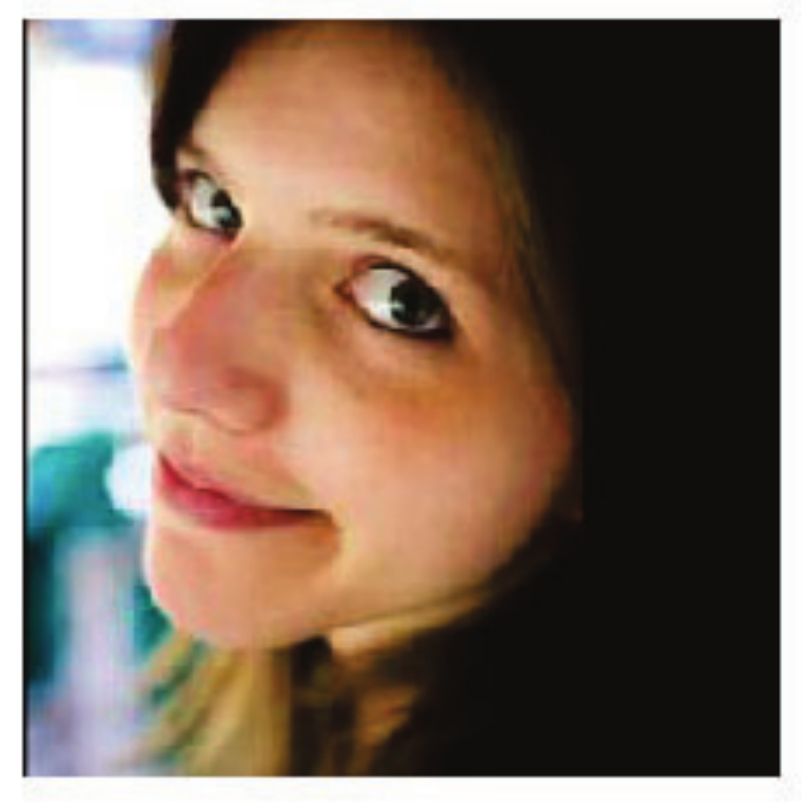}
		}
		\subfigure[]{
			\includegraphics[width=0.155\linewidth]{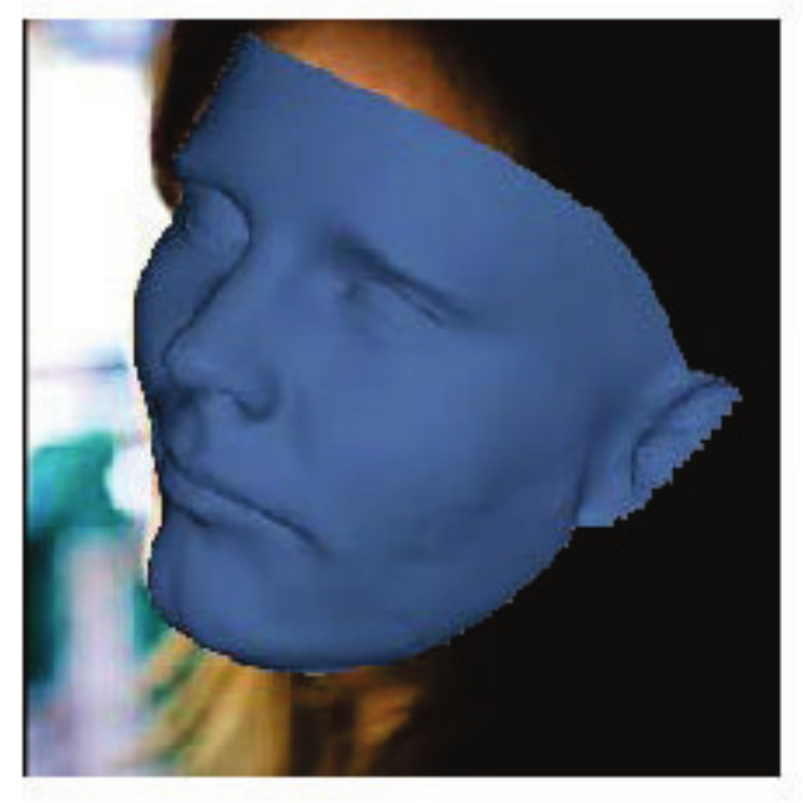}
		}
		\subfigure[]{
			\includegraphics[width=0.155\linewidth]{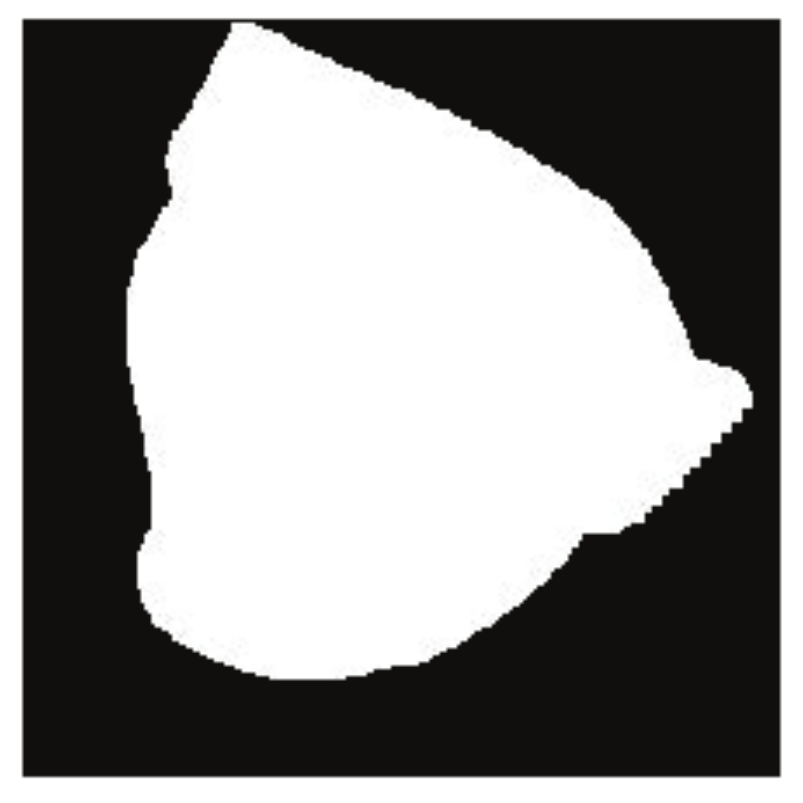}
		}
		\subfigure[]{
			\includegraphics[width=0.155\linewidth]{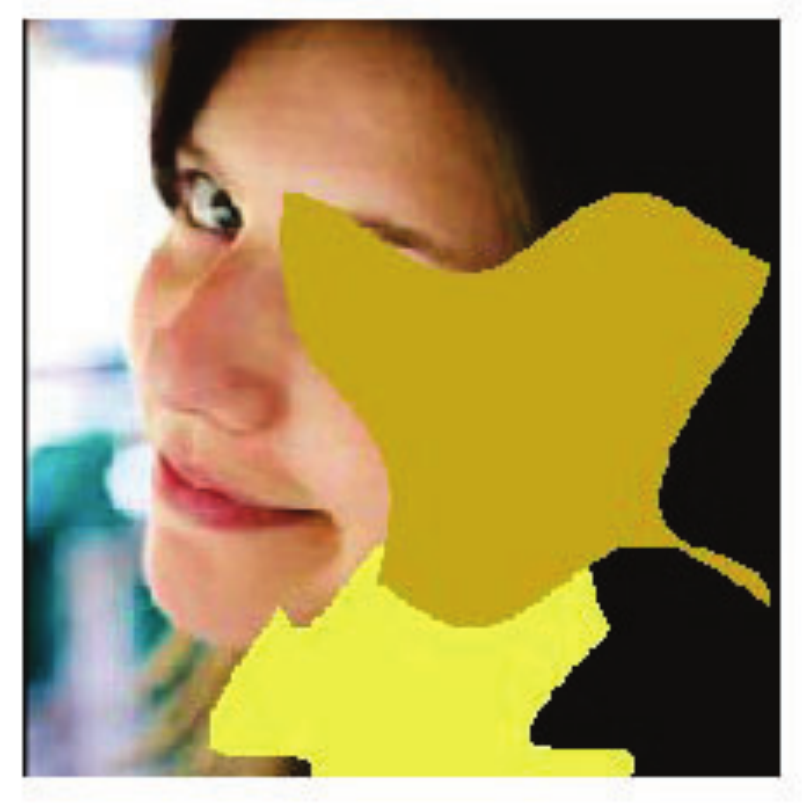}
		}
		\subfigure[]{
			\includegraphics[width=0.155\linewidth]{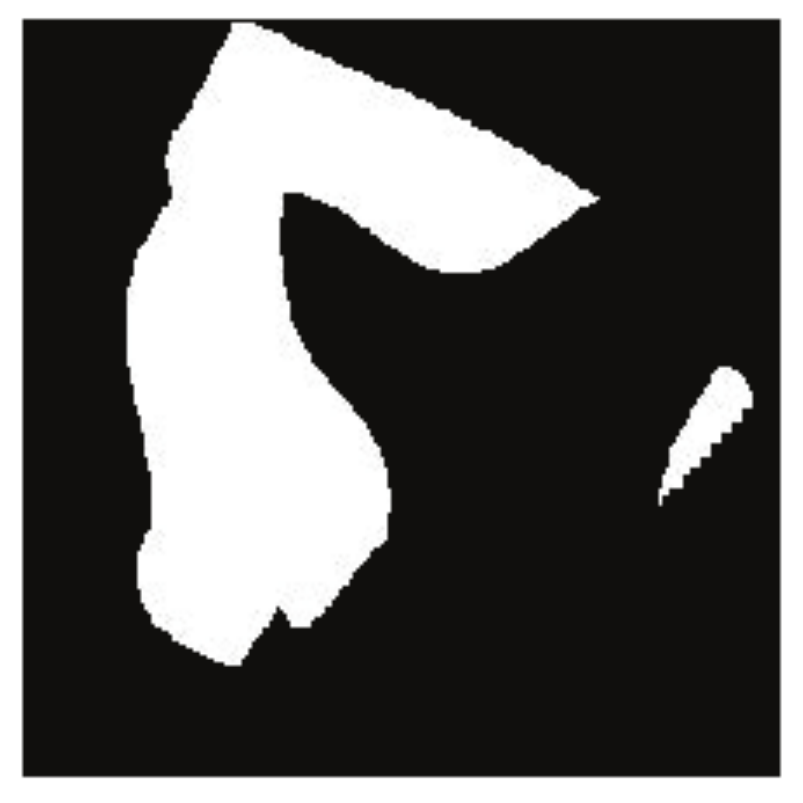}
		}
		
		\caption{An example of the synthetic attention mask annotation. (a) is the original face image. (b) is the ground truth face geometry projected on the image plane. (c) is the binary image of the projected face. (d) is the face image augmented with synthetic occlusions. (e) is the ground truth attention mask corresponding to (d).}
		\label{fig:synth}
		
	\end{figure}
	
	To deal with the occlusions, we adopt an attention mechanism to extract features mainly from the visible face regions in the input images. This sub-network is a side branch consisting of five convolutional layers. It outputs a soft attention mask assigning high values to the pixels corresponding to the visible regions and low values to the pixels in the occluded regions and background region. The attention is applied to the low-level features to make a trade-off between resolution and accuracy.
	
	Considering the attention mask may be inaccurate in some cases, we do not discard the information of the regions with low attention entirely. We instead highlight the features of the visible face regions according to the attention values. This operation can be formulated as
	\begin{equation}
		\mathbf{F_a}=\mathbf{F_l}\odot\exp(\mathbf{A}),
	\end{equation}
	where $\mathbf{F_a}$ denotes the obtained weighted feature map, $\mathbf{F_l}$ is the low-level feature map, and $\mathbf{A}$ is the attention mask.
	
	Once we get the attention mask, we are able to estimate the visibility of the vertices in the pose-dependent face through
	\begin{equation}
		\mathbf{Vis}(i)=\begin{cases}
			
			0&\mbox{if $n_z<0$ }\\
			
			\mathbf{A}(\lfloor x_i \rfloor,\lfloor y_i \rfloor)&\mbox{if $n_z\geq0$},
			
		\end{cases}
	\end{equation}
	where the 3D position of vertex $i$ is $(x_i,y_i,z_i)$. The normal direction of vertex $i$ that can be obtained from its adjacent vertices is $(n_x,n_y,n_z)$. The vertex visibility is employed in the self-alignment module and is discussed detailedly in Sec.~\ref{sec:svd}.
	
\begin{figure*}[tbp]
		\centering
		
		\includegraphics[width=1.0\linewidth]{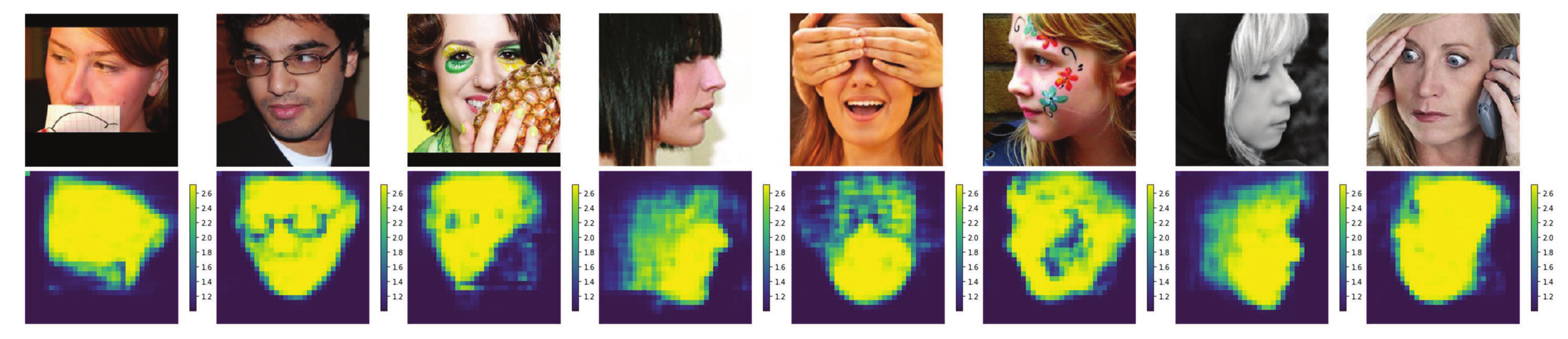}
		
		\caption{The top row shows the input images and the bottom row shows the attention masks predicted by our attention branch.}
		\label{fig:occlusion}
	\end{figure*}

	Since no database has the ground truth for face occlusion annotation, we simulate occlusions through data augmentation. Specifically, we project the 3D face geometry to the image plane to generate a binary map that indicates the full face region. Then we overlay patterns with random shapes to the input image and use them as the occlusions as shown in Fig.~\ref{fig:synth}. We use the resulting binary image as the ground truth of the attention mask. In this way, we can obtain the ground truth data for the training of the attention branch. Although occlusions in the real world can be very diverse, our experiments find the proposed method has the ability to predict real-world occlusions as shown in Fig.~\ref{fig:occlusion}.

	\subsection{Self-Alignment Module}
	\label{sec:svd}
	
	This module aims to extract the pose information from the pose-dependent face $\mathbf{P}$ and the shape information from the pose-independent face $\mathbf{S}$ (see Fig.~\ref{fig:framework} for the two faces $\mathbf{P}$ and $\mathbf{S}$). It is achieved by estimating the similarity transformation matrices between $\mathbf{P}$ and $\mathbf{S}$ (i.e., $f$, $\mathbf{R}$, and $\mathbf{t}$ in Eq.~\ref{eq_1}).
	The similarity transformation matrix is estimated using two sets of correspondent landmarks, $\mathbf{K_S}$ and $\mathbf{K_P}$, extracted from $\mathbf{S}$ and $\mathbf{P}$, respectively.
	The pixels with the same coordinates in the UV maps of $\mathbf{S}$ and $\mathbf{P}$ are
	semantically correspondent (e.g., the nose tip corresponds to two locations with the same coordinates in the UV maps of $\mathbf{S}$ and $\mathbf{P}$). Thus, in our method, $\mathbf{K_S}$ and $\mathbf{K_P}$ are extracted based on the same UV coordinates set. Specifically, we extract the 68 landmarks as the same as~\cite{3DDFA} for $\mathbf{K_S}$ and $\mathbf{K_P}$.
	
	We do not simply use all the 68 landmarks. We propose to use more reliable facial vertices, i.e., the visible landmarks, for the alignment. We define a diagonal matrix $\mathbf{W}\in\mathbb{R}^{k\times k}$ that represents the weight of each landmark with the visibility. The weight of the $i$th landmark is formulated as
	\begin{equation}
		\mathbf{W}(i,i)=\mathbf{Vis}(i) + eps,
	\end{equation}
	where $\mathbf{Vis}(i)$ denotes the visibility of the $i$th landmark as mentioned earlier. $eps$ is set to $0.1$ in our implementation to avoid the divide-by-zero error caused by the estimated visibility of all landmarks being zero.
	The estimation of the similarity transformation goes as the following steps: we first estimate
	$f$, and then estimate $\mathbf{R}$ and $\mathbf{t}$ using the singular value decomposition method.
	Specifically,
	We first compute the two weighted centroids $\mathbf{M_S}$ and $\mathbf{M_P}$ of the two sets of landmarks $\mathbf{K_S}$ and $\mathbf{K_P}$ by
	\begin{equation}
		\mathbf{M_S}=\frac{\sum_{i=1}^{k}\mathbf{W}(i,i)*\mathbf{K_S}(i)}{|\mathbf{W}|},
	\end{equation}
	\begin{equation}
		\mathbf{M_P}=\frac{\sum_{i=1}^{k}\mathbf{W}(i,i)*\mathbf{K_P}(i)}{|\mathbf{W}|}.
	\end{equation}
	Then we obtain $f$ by
	\begin{equation}
		f=\frac{\sum_{i=1}^{k}\lVert \mathbf{K_P}(i)-\mathbf{M_P} \rVert }{\sum_{i=1}^{k}\lVert \mathbf{K_S}(i)-\mathbf{M_S} \rVert}.
	\end{equation}
	After that, we normalize $\mathbf{K_S}$ and $\mathbf{K_P}$ as bellow:
	\begin{equation}
		\mathbf{K_S^{'}}=f*(\mathbf{K_S}-\mathbf{M_S}),
	\end{equation}
	\begin{equation}
		\mathbf{K_P^{'}}=\mathbf{K_P}-\mathbf{M_P}.
	\end{equation}
	Performing SVD to $\mathbf{H}=\mathbf{K_S^{'}*\mathbf{W}*\mathbf{K_P^{'\mathrm{T}}}} $,
	we have $[U,\Sigma,P]=\mathrm{SVD}(\mathbf{H})$, the rotation matrix $\mathbf{R}$ between $\mathbf{K_S^{'}}$ and $\mathbf{K_P^{'}}$ can be obtained by
	\begin{equation}
		\mathbf{R}=P*U^\mathrm{T},
	\end{equation}
	and $\mathbf{t}$ can be obtained by
	\begin{equation}
		\mathbf{t}=\mathbf{M_P}-\mathbf{R}*\mathbf{M_S}.
	\end{equation}
	The pose estimation from landmarks requires only a small amount of computation and has high accuracy because the estimation is performed on a number of reliable landmarks on the two regressed faces.

	\subsection{Loss Functions}
	\label{sec:loss}
	The loss of our SADRNet consists of 6 components, the face geometry loss $\mathcal{L}_G$, the deformation loss $\mathcal{L}_D$, the pose-dependent face loss $\mathcal{L}_P$, the attention mask loss $\mathcal{L}_A$, the edge length loss $\mathcal{L}_E$, and the normal vector loss $\mathcal{L}_V$.
	Since the significance of the vertices in different face regions may differ considerably, we adopt a weight mask $\mathbf{M}$ as~\cite{PRN} to $\mathcal{L}_D$, $\mathcal{L}_P$, and $\mathcal{L}_G$. We adjust the weight ratio of the four sub-regions for better training results. We use 16:12:3:0 for sub-region1 (the landmarks): sub-region2 (the eyes, nose, and mouth): sub-region3 (the cheek, chin and forehead): sub-region4 (the neck).
	
	$\mathcal{L}_G, \mathcal{L}_D$, and $\mathcal{L}_P$ can be obtained by the weighted average Euclidean distance between the estimated value and the ground truth, i.e., generalized as
	\begin{equation}
		\mathcal{L}_N=\sum_{u=1}^{h}\sum_{v=1}^{w}\lVert \mathbf{N}(u,v)-\mathbf{\hat{N}}(u,v) \rVert_2\cdot \mathbf{M}(u,v).
	\end{equation}
	$N$ denotes any one of the face geometry $\mathbf{G}$, the pose-dependent face $\mathbf{P}$, and the shape deformation $\mathbf{D}$. $h$ and $w$ are the height and width of the maps. $\mathbf{N}(u,v)$ denotes the estimated 3D coordinates of the vertex at the UV location $(u,v)$ in the UV maps of $\mathbf{G}$,  $\mathbf{P}$, or $\mathbf{D}$. The symbol $\mathbf{\hat{~}}$ denotes the corresponding ground truth. $\mathbf{M}(u,v)$ denotes the weight value at the pixel location $(u,v)$ in $\mathbf{M}$.
	
	We use binary cross entropy (BCE) between the predicted attention mask $\mathbf{A}$ and the ground truth $\mathbf{\hat{A}}$ to compute $\mathcal{L}_A$.
	
	The edge length loss $\mathcal{L}_E$ is defined based on the lengths of the edges in the pose-independent face $\mathbf{S}$ as
	\begin{equation}
		\mathbf{E}_{ij}= \mathbf{S}(u_i,v_i)-\mathbf{S}(u_j,v_j)  ,
	\end{equation}
	\begin{equation}
		\mathcal{L}_E=\sum_{(i,j)\in \mathcal{E}}^{} | \lVert\mathbf{E}_{ij}\rVert_2-\lVert\mathbf{\hat{E}}_{ij}\rVert_2 |.
	\end{equation}
	$\mathcal{E}$ is the edges set. It is composed of every pair of adjacent vertices $(i,j)$ in the UV map. $(u_i,v_i)$ is the UV coordinates of vertex $i$. $\mathbf{E}_{ij}$ is the edge vector.
	
	The normal vector loss $\mathcal{L}_V$ is defined as
	\begin{equation}
		\mathbf{n}_{ijk}=\frac{\mathbf{E}_{ij}\times\mathbf{E}_{jk}}{\lVert \mathbf{E}_{ij}\times\mathbf{E}_{jk} \rVert_2},
	\end{equation}
	
	\begin{equation}
		\begin{aligned}
			\mathcal{L}_V =\sum_{(i,j,k)\in \mathcal{T}}&\left|\left \langle \frac{\mathbf{E}_{ij}}{\lVert\mathbf{E}_{ij}\rVert_2}, \mathbf{n}_{ijk} \right \rangle \right| \\
			+&\left|  \left \langle \frac{\mathbf{E}_{jk}}{\lVert\mathbf{E}_{jk}\rVert_2}, \mathbf{n}_{ijk} \right \rangle\right|                             \\
			+&\left|\left \langle \frac{\mathbf{E}_{ki}}{\lVert\mathbf{E}_{ki}\rVert_2}, \mathbf{n}_{ijk} \right \rangle \right|,
		\end{aligned}
	\end{equation}
	where $\mathcal{T}$ is the triangle facets set, $\mathbf{n}_{i,j,k}$ is the normal vector of the triangle facet $(i,j,k)$. The edge length loss and the normal vector loss are used to generate better-looking face models.

	The entire loss function of SADRNet is given by
	\begin{equation}
		\mathcal{L}=\beta_G\mathcal{L}_G+\beta_D\mathcal{L}_D+\beta_P\mathcal{L}_P+\beta_A\mathcal{L}_A+\beta_E\mathcal{L}_E+\beta_V\mathcal{L}_V,
	\end{equation}
	where $\beta_G$, $\beta_D$, $\beta_P$, $\beta_A$, $\beta_E$, $\beta_V$ are respectively set to 0.1, 0.5, 1, 0.05, 1, 0.1 in our implementation.
	
	\subsection{Implementation Details}
	\label{sec:detail}
	Our network takes cropped-out $256\times256\times3$ images as the input, regardless of the effect of the face detector. The network starts with a single convolution layer followed by a low-level feature extractor that consists of 6 residual blocks~\cite{resnet_2016_CVPR} and outputs a $32\times32\times128$ feature map. The attention sub-network contains 5 convolution layers and a sigmoid activation. The high-level feature extractor contains 4 residual blocks and outputs a $8\times8\times512$ feature map. The decoder starts with 10 transpose convolution layers that up-sample the feature map to $64\times64\times64$, followed by 7 transpose convolution layers to output a UV map of the face shape with size $256\times256\times3$ and another 7 layers to output a UV map of the pose-dependent face with size $256\times256\times3$.

	We follow \cite{3DDFA,PRN,MMFace2019_CVPR} and use the full \textit{300W-LP}~\cite{3DDFA} as the training set. 300W-LP contains $122,450$ face images generated from \textit{300W}~\cite{3DDFA} by 3D rotation around the y axis and horizontal flipping. Since there is no sample with a larger than 90 degrees yaw angle in 300W-LP, we generate 5,000 samples with yaw angles range from 90 to 105 degrees by 3D rotation to supplement the dataset.  Similar to~\cite{PRN}, we augment the training data by randomly rotating the image from -90 to 90 degrees, translating in the range of 0 to 10 percent of input size, scaling the image size from 0.9 to 1.1, and scaling the color channel separately from 0.6 to 1.4. We also generate synthetic occlusions as explained in Sec.~\ref{subsec:attention}. We use Adam optimizer with a gradual warm-up strategy~\cite{gradual_warmup}. We start from a learning rate of 1e-5 and increase it by a constant amount at each iteration. After 4 epochs, the learning rate reaches 1e-4. Then we use an exponential scheduler that decays the learning rate by 0.85 every epoch. The batch size is set to 16. We train our network for 25 epochs.
	
	The original annotations of 300W-LP are 3DMM parameters, so the ground truth of $\mathbf{D}$ can be generated by computing $\mathbf{D}=\sum_{i}\alpha_i\mathbf{A}_i$, where $(\alpha_1,\alpha_2,\dots)$ are the shape parameters, $(\mathbf{A}_1,\mathbf{A}_2,\dots)$ are the shape basis vectors (including the identity and expression basis vectors).

	\section{Experiments}
\begin{figure*}[tbp]
		
		\begin{center}
			
			\includegraphics[width=0.95\linewidth]{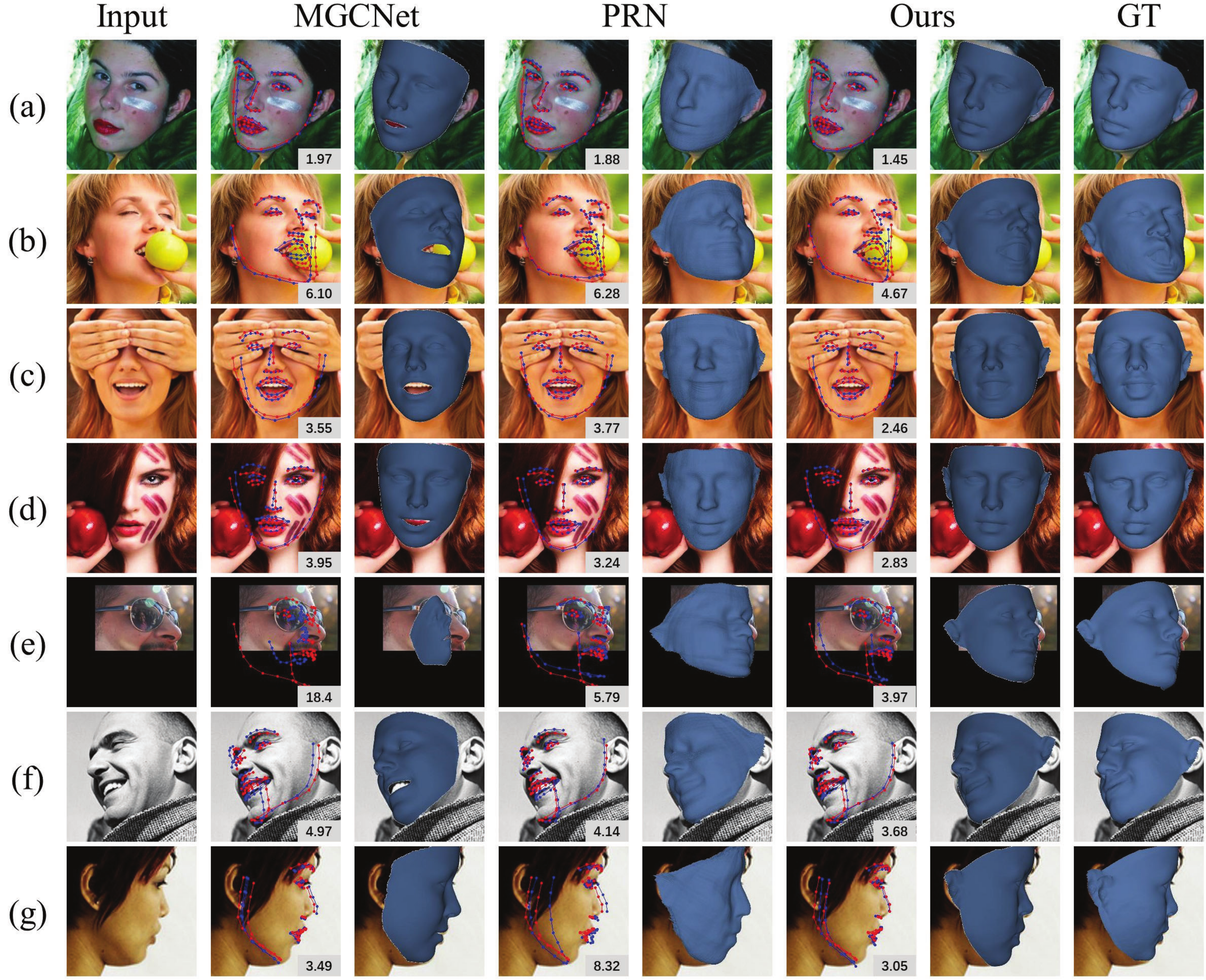}
		\end{center}
		
		\caption{The qualitative comparison on AFLW2000-3D dataset. The estimated landmarks are in blue and the ground truth landmarks are in red. NME(\%) is shown at the bottom right of each result.}
		\label{fig:all}
	\end{figure*}

	In this section, we evaluate the performance of the proposed method in terms of 3D face reconstruction, dense face alignment and head pose estimation. Our SADRNet is quantitatively compared with state-of-the-art methods including CMD (2019~\cite{FPS}), SPDT (2019~\cite{semi_2019_ICCV}), PRN (2018~\cite{PRN}) and 3DDFAv2 (2020~\cite{3ddfav2}).
	The qualitative comparison of PRN, MGCNet~\cite{MGC} and our method are demonstrated in Fig.~\ref{fig:all}.
	
	\begin{table*}[tbp]
		\caption{Performance comparison on AFLW2000-3D on the task of sparse alignment (68 landmarks) and dense alignment (45K points).  The NME (\%) are reported. Images with different yaw angle ranges are also evaluated separately. ``balanced'' denotes the results on the subset with balanced distribution of the yaw angles. ``mean'' denotes the average results on the entire dataset. The best results are highlighted in bold.}
		\begin{center}
			\begin{tabular}{|c|c|c|c|c|c|c|c|c|}
				\hline
				\multirow{3}*{Method}       &                       \multicolumn{6}{c|}{68 points}                        & \multicolumn{2}{c|}{45k points} \\ \cline{2-9}
				&                    \multicolumn{5}{c|}{2D}                     &     3D     &    {2D}    &        {3D}        \\ \cline{2-9}
				& {0 to 30}  & {30 to 60} & {60 to 90} &  Balanced  &    Mean    &    Mean    &    Mean    &        Mean        \\ \hline
				3DDFA~\cite{3DDFA}         &    3.78    &    4.54    &    7.93    &    5.42    &    6.03    &    7.50    &    5.06    &        6.55        \\
				3DFAN~\cite{how_far}        &    2.77    &    3.48    &    4.61    &    3.62    &     -      &     -      &     -      &         -          \\
				DeFA~\cite{defa_2017_ICCV}     &     -      &     -      &     -      &    4.50    &    4.36    &    6.23    &    4.44    &        6.04        \\
				3DSTN~\cite{3dstn}         &    3.15    &    4.33    &    5.98    &    4.49    &     -      &     -      &     -      &         -          \\
				Nonlinear 3DMM~\cite{on_learning} &     -      &     -      &     -      &    4.12    &     -      &     -      &     -      &         -          \\
				
				PRN~\cite{PRN}           &    2.75    &    3.51    &    4.61    &    3.62    &    3.26    &    4.70    &    3.17    &        4.40        \\
				DAMDN~\cite{dual_attention}    &    2.90    &    3.83    &    4.95    &    3.89    &     -      &     -      &     -      &         -          \\
				CMD~\cite{FPS}           &     -      &     -      &     -      &    3.90    &     -      &     -      &     -      &         -          \\
				SPDT~\cite{semi_2019_ICCV}     &    3.56    &    4.06    & {\bf 4.11} &    3.88    &     -      &     -      &     -      &         -          \\
				3DDFAv2~\cite{3ddfav2}      &   \bf{2.63}    &    3.42    &    4.48    &    3.51    &     -      &     -      &     -      &         4.18          \\
				{\bf SADRNet (ours)}         & 2.66 & {\bf 3.30} &    4.42    &    {\bf 3.46}    & {\bf 3.05} & {\bf 4.33} &    {\bf 2.93}    &        {\bf4.02}        \\
				\hline
			\end{tabular}
			
		\end{center}
		\label{tab:sparse alignment}
	\end{table*}
	
	\subsection{Evaluation Datasets}
	{\bf AFLW2000-3D}~\cite{3DDFA} is an in-the-wild dataset with large variations in pose, illumination, expression, and occlusion. There are 2,000 images annotated with 68 3D landmarks and fitted 3DMM parameters to recover the ground truth face model. We evaluate both face alignment and 3D face reconstruction performance on this dataset.
	
	{\bf Florence}~\cite{florence} is a publicly available database of 53 subjects. The ground truth annotations are meshes scanned by a structured-light system. Similar to~\cite{VRN,PRN,MMFace2019_CVPR}, each face mesh is rendered in 20 poses: a pitch angle of -15, 0, 20, or 25 degrees and a yaw angle of -80, -40, 0, 40, and 80 degrees to generate the face images. We evaluate the 3D face reconstruction performance on this dataset.
	
	\subsection{Face Alignment}
	We employ normalized mean error (NME) as the evaluation metric. It is defined as the normalized mean Euclidean distance between each pair of corresponding points in the predicted result $\mathbf{p}$ and the ground truth $\mathbf{\hat{p}}$:
	\begin{equation}
		\mathrm{NME}=\frac{1}{N}\sum_{i=1}^{N}\frac{\lVert \mathbf{p_i} - \mathbf{\hat{p}_i} \rVert_2}{d}.
	\end{equation}
	For a fair comparison, the normalization factor $d$ of the compared methods is computed in the same way. For sparse and dense alignment, the normalization factor of NME is defined as $\sqrt{h*w}$ where $h$ and $w$ are the height and width of the bounding box of all the evaluated points.
	
	We evaluate the performance of 2D and 3D sparse alignment on the point set of 68 landmarks. We evaluate the performance of 2D and 3D dense alignment on the point set of around 45K points selected from the largest common face region of different methods as in ~\cite{PRN}. Following the settings of previous works~\cite{PRN,3DDFA,FPS} on 2D sparse alignment, we also test the images with yaw angles in $[0^\circ,30^\circ)$ (1,306 samples), $[30^\circ,60^\circ)$ (462 samples) and $[60^\circ,90^\circ]$ (232 samples) separately, as well as the balanced subsets of each angle interval.

	\begin{figure}[hpt]
		\centering
		\includegraphics[width=1.0\linewidth]{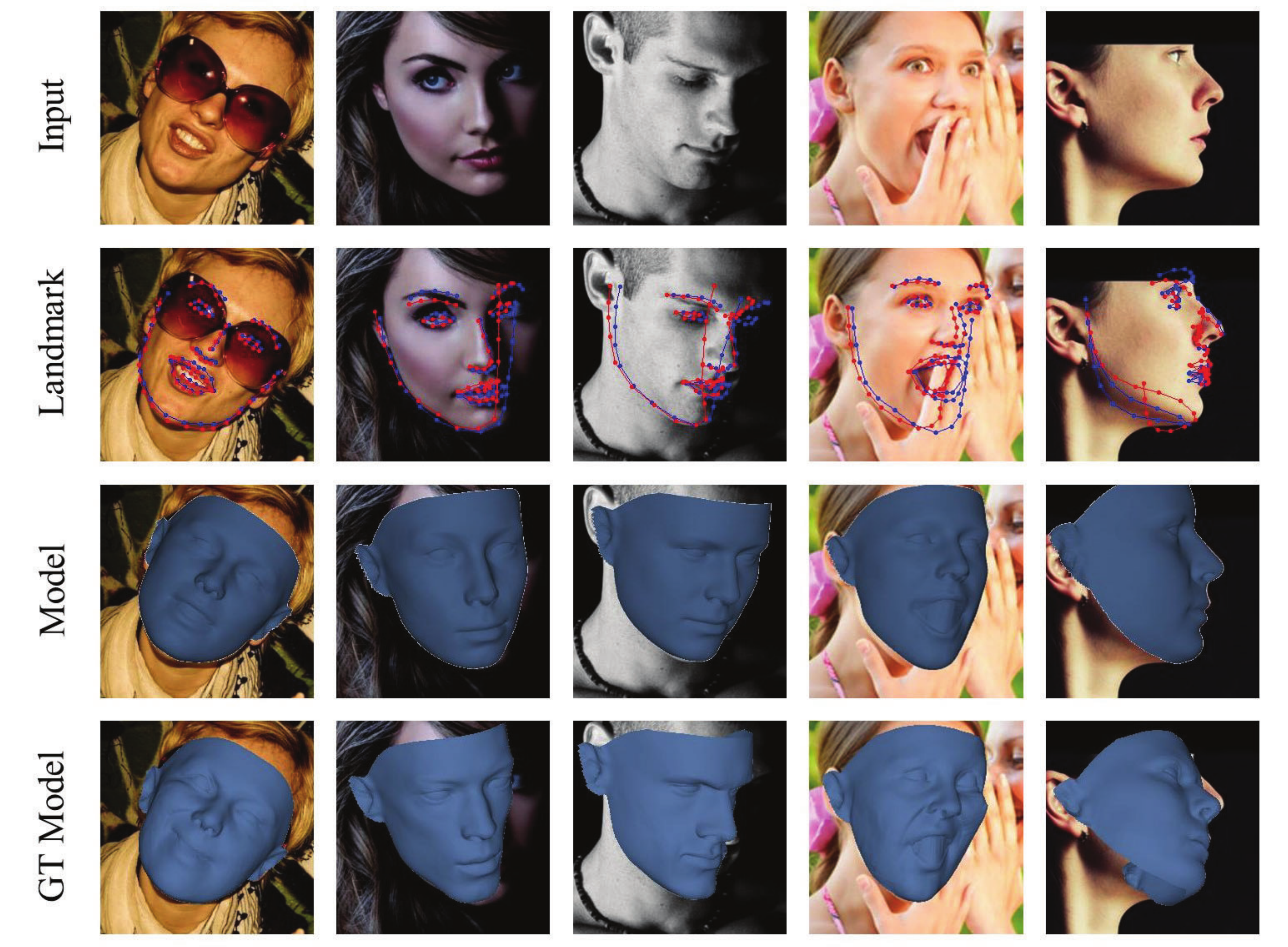}
		\caption{ Results on AFLW2000-3D from our SADRNet. Our results are more accurate than the ground truth. From the top row to the bottom row are the input images, the sparse alignment results of SADRNet and the corresponding ground truth (blue for our method and red for the ground truth), the reconstructed face models, and the ground truth face models.}
		\label{fig:fail}
	\end{figure}
	
	The quantitative results on AFLW2000-3D are shown in Table~\ref{tab:sparse alignment}. The results of other methods are from published papers or produced by the official open-source codes.
	We can see our method is superior to the other methods on most metrics. Especially on 3D and dense alignment tasks, our method has taken a clear lead.	
	SPDT~\cite{semi_2019_ICCV} generates rendered large pose training samples and uses a CycleGAN~\cite{cycle_gan_2017_ICCV} to transform the rendering style. The large pose samples in our training database 300W-LP are mostly obtained by 3D image rotation with large distortion. We argue that they outperform us for sparse alignment in-between 60 and 90 degrees is because of the quality gap of training data.
	
	Visualized results of challenging samples in AFLW2000-3D are demonstrated in Fig.~\ref{fig:all}, in which various degrees of occlusion and different face orientations are evaluated. We compare our method with two representative methods: MGCNet~\cite{MGC} and PRN~\cite{PRN}. MGCNet is a model-based method that fits the shape and the pose parameters of 3DMM by CNN. PRN is a model-free method that directly infers 3D coordinates of face mesh vertices with the UV position map.
	Our pose estimation method based on the alignment using visibility is more accurate and robust than previous works.
	
	On the samples where the faces are partially occluded, as shown in Fig.~\ref{fig:all}c and Fig.~\ref{fig:all}d,
	the inaccuracy of the estimated poses of MGCNet leads to misalignment. And under severe occlusion, e.g., Fig.~\ref{fig:all}e, MGCNet fails to obtain a reasonable result, while our method is still able to precisely estimate the head pose. PRN jointly obtains pose and shape and works well for the visible facial region. However, PRN tends to give results with larger errors for invisible regions (see the misaligned landmarks in Fig.~\ref{fig:all}b and Fig.~\ref{fig:all}g).
	It is worth noting that AFLW2000-3D~\cite{3DDFA} is semi-automatically annotated and has some
	inaccurate annotations in some cases.
	In Fig.~\ref{fig:fail}, we demonstrate some examples from AFLW2000-3D that our predictions have relatively larger NME but are apparently more accurate than the ground truth. This may narrow our method's superiority margin.

	\begin{table}[tbp]
		\caption{Performance comparison on 3D face reconstruction. Evaluations are conducted on the AFLW2000-3D dataset and Florence dataset. Around 45K points are used on AFLW2000-3D and 19K points are used on Florence.  The NME (\%) normalized by outer interocular distance are reported. The best results are highlighted in bold.}
		\begin{center}
			\begin{tabular}{|c|c|c|c|c|c|}
				\hline
				\multirow{2}{*}{Method}   &         \multicolumn{4}{c|}{AFLW2000-3D}         &  Florence  \\ \cline{2-6}
				&   0 ~ 30   &  30 ~ 60   &  60 ~ 90  &    Mean    &    Mean    \\ \hline
				3DDFA~\cite{3DDFA}     &     -      &     -      &     -     &    5.36    &    6.38    \\
				DeFA~\cite{defa_2017_ICCV} &     -      &     -      &     -     &    5.64    &     -      \\
				VRN - Guided~\cite{VRN}   &     -      &     -      &     -     &     -      &    5.26    \\
				PRN~\cite{PRN}       &    3.72    &    4.04    &   4.45    &    3.96    &    3.75    \\
				SPDT~\cite{semi_2019_ICCV} &     -      &     -      &     -     &    3.70    &    3.83    \\
				3DDFAv2~\cite{3ddfav2}   &     -      &     -      &     -     &     -      &    3.56    \\
				SADRNet (ours)       & {\bf 3.17} & {\bf 3.42} & {\bf3.36} & {\bf 3.25} & {\bf 3.12} \\ \hline
			\end{tabular}
			
		\end{center}
		\label{tab:reconstruction}
	\end{table}

	\subsection{3D Face Reconstruction}
	On this task, we use NME normalized by 3D outer interocular distance as the evaluation metric. We follow the settings of ~\cite{PRN} to evaluate our method on AFLW2000-3D~\cite{3DDFA} and Florence~\cite{florence}.
	For AFLW2000-3D, we use the same set of points as we do for dense face alignment. The results under different yaw angles are also reported. As for Florence, we choose the face region containing 19K points. We use iterative closest point (ICP) to align the results to the corresponding ground truth as in ~\cite{PRN}. Table~\ref{tab:reconstruction} gives the quantitative comparison.
	Our method is robust to pose variations and achieves around 13\% improvement over the state-of-the-art method on both datasets.

	As our method employs a nonlinear shape deformation decoder that is more powerful than the linear bases based representation of MGCNet, our network can reconstruct more shape variations. It is worth mentioning that the output space of PRN is also nonlinear, but its large space, caused by entangled shape and pose, increases the learning difficulty for the high-frequency details. It is shown in Fig.~\ref{fig:all} that eyes, lips, and noses of the face models reconstructed by our method have more details than PRN. Our attention-aware mechanism also contributes to the robustness toward the occlusions. In Figs.~\ref{fig:all}b-e, our reconstructed faces maintain natural appearances while those from PRN are distorted in occluded areas.

	\begin{table}[tbp]
		\caption{Performance comparison on head pose estimation. Evaluations are conducted on the AFLW2000-3D dataset. The MAEs of head pose parameters are reported. The best results are highlighted in bold.}
		\begin{center}
			\begin{tabular}{|c|c|c|c|c|}
				\hline
				\multirow{2}{*}{Method} &         \multicolumn{4}{c|}{AFLW2000-3D}         \\ \cline{2-5}
				&    yaw     &   pitch    &   roll    &    Mean    \\ \hline
				FSANet~\cite{FSA}    &    4.50    &    6.08    &   4.64    &    5.07    \\
				GLDL~\cite{GLDL}     &    3.02    &    5.06    &   3.68    &    3.92    \\
				QuatNet~\cite{Quatnet}  &    3.97    &    5.61    &   3.92    &    4.50    \\
				FDN~\cite{FDN}      &    3.78    &    5.61    &   3.88    &    4.42    \\
				MNN~\cite{MNN}      &    3.34    &    4.69    &   3.48    &    3.83    \\ \hline
				SADRNet (ours)      &    2.93    &    5.00    &   3.54    &    3.82    \\
				SADRNet-fix (ours)    & {\bf 2.93} & {\bf 4.43} & {\bf2.95} & {\bf 3.44} \\ \hline
			\end{tabular}
			
		\end{center}
		\label{tab:pose estimation}
	\end{table}

	\subsection{Head Pose Estimation}
	On this task, we use MAE (mean absolute error) of the head pose parameters as the evaluation metric:
	\begin{equation}
		\mathrm{MAE}=\frac{1}{N}\sum_{i=1}^{N}\left|\mathbf{p}_i-\mathbf{\hat{p}}_i \right|,
	\end{equation}
	where $\mathbf{p}$ represents the predicted pose parameters and $\mathbf{\hat{p}}$ represents the ground truth.
	
\begin{figure}[hpt]
		\centering
		\includegraphics[width=1\linewidth]{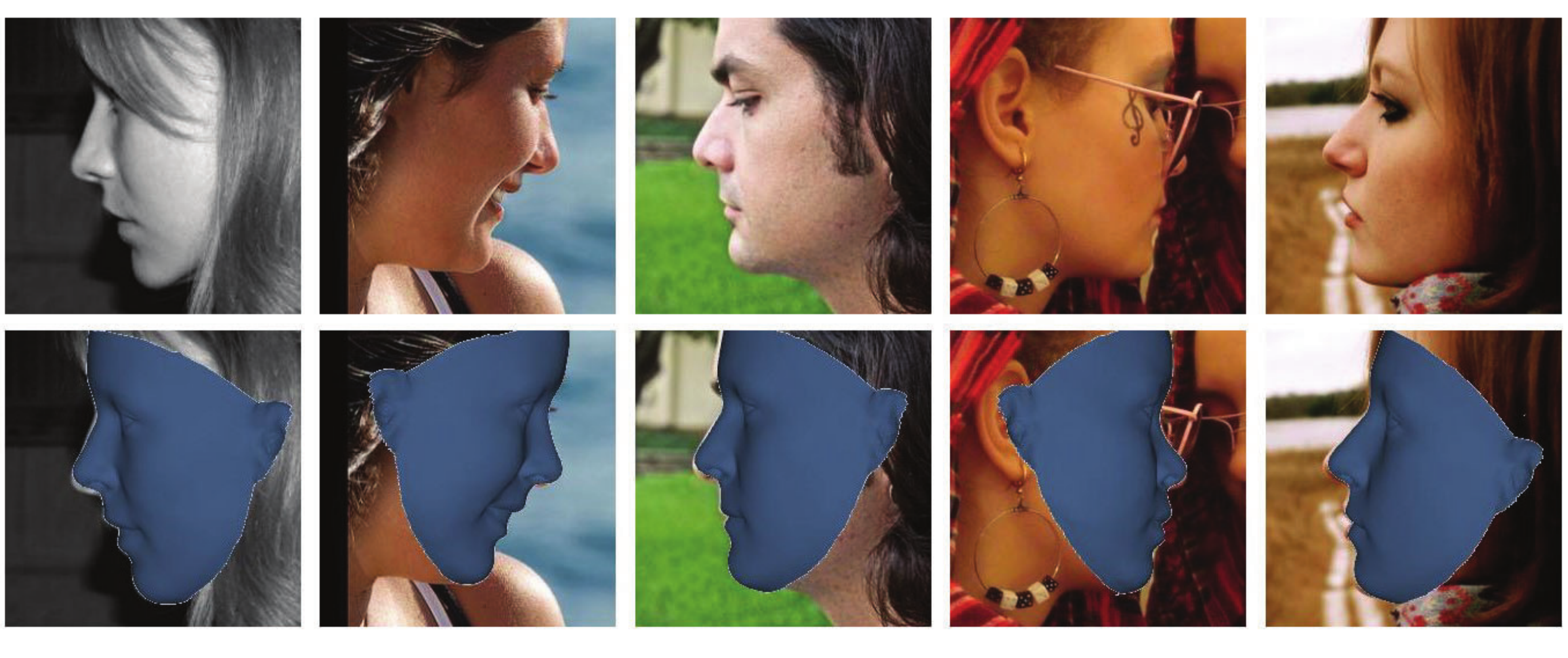}
		\caption{Some samples in AFLW2000-3D, in which our estimated pitch angle and roll angle have an error of more than 20 degrees, while the yaw angle errors are less than 5 degrees.}
		\label{fig:gimbal}
	\end{figure}

	In Table~\ref{tab:pose estimation}, we compare our proposed self-aligned module with the SOTA head pose estimation methods. The actual output of our method is the transformation matrix. So we convert the rotation matrix to Euler angles to calculate MAE with the ground truth. However, when the yaw angle is close to 90 degrees, the angle conversion suffers a serious gimbal lock problem. Some examples are shown in Fig.\ref{fig:gimbal}. Their yaw angles' errors are low, and the faces' orientations seem nice, but the errors of the pitch and roll angles are extremely high. It will make the quantitative evaluation result worse than the actual performance. Thus, besides the initial evaluation results, we also report a result marked as SADRNet-fix that ignores the samples with more than 20 degrees pitch or roll error and less than 5 degrees yaw error.
	However, the negatively biased performance of our method (SADRNet) is still equivalent to the best standalone head pose estimation method.
	
	\subsection{Ablation Study}
	\begin{figure*}[hpt]
		\centering
		\includegraphics[width=1\linewidth]{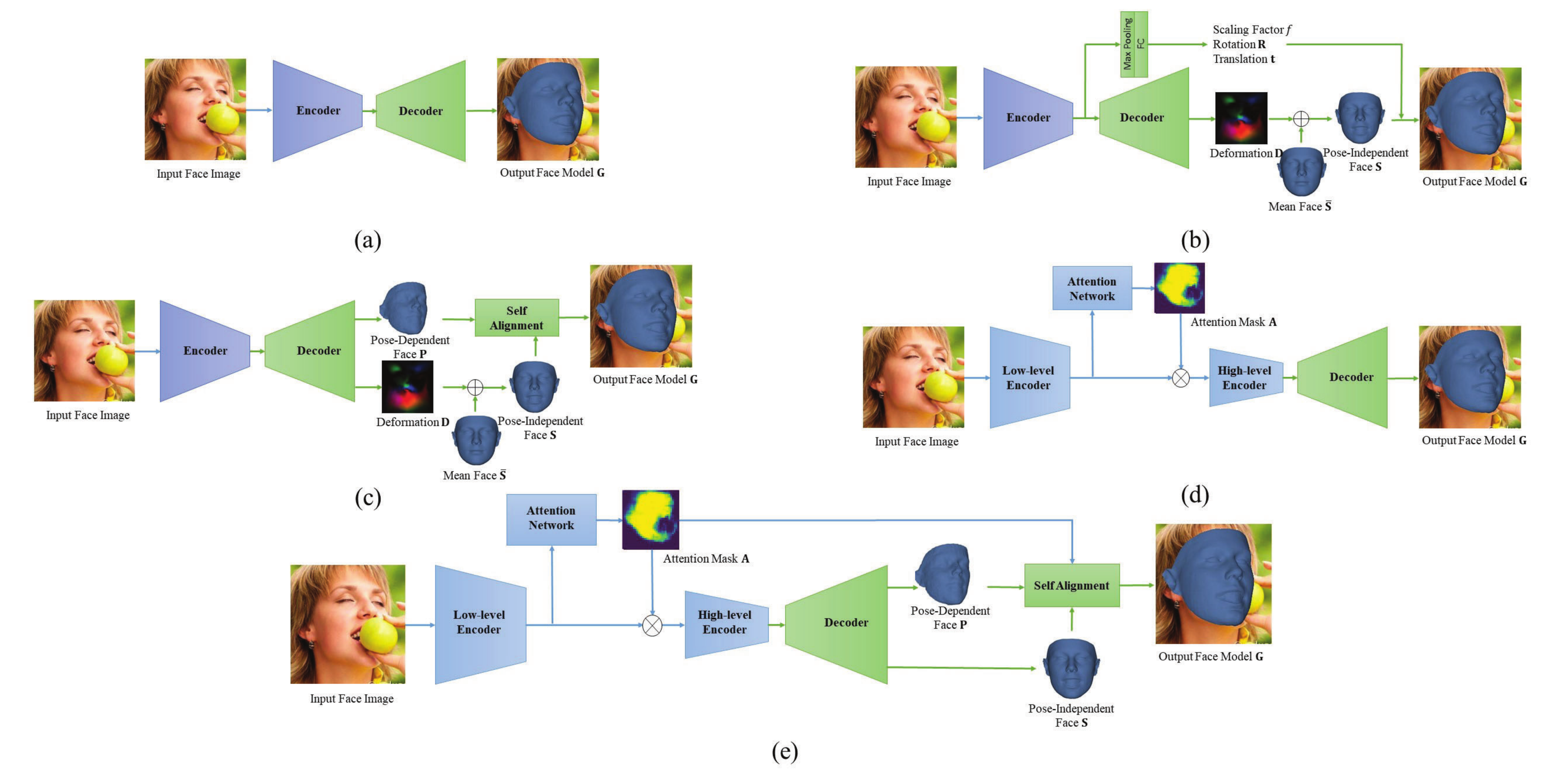}
		\caption{Alternative frameworks of baselines in ablation study. The figures (a)-(e) correspond to the 1st-5th result rows in Table~\ref{tab:ablation}, respectively.}
		\label{fig:ablation_framework}
	\end{figure*}
	We conduct ablation experiments on AFLW2000-3D to analyze the effectiveness of the following three modules: 1) the attention mechanism, 2) the self-alignment module, and 3) the regression via the shape deformation. The experiments are conducted on three
	tasks: sparse alignment, dense alignment, and 3D face reconstruction. The results are summarized in Table~\ref{tab:ablation}.
	
	\begin{table}[tbp]
		\caption{Ablation study on AFLW2000-3D. The ``AT'' column indicates using the attention mechanism or not. The ``SA'' column indicates using the self alignment module or not. The ``DF'' column indicates
			whether 3D-DFAFR is achieved by regressing shape deformation ``$\mathbf{D}$'' or not.}
		\label{tab:ablation}
		\centering
		\begin{tabular}{|c|c|c|c|c|c|c|c|}
			\hline
			\multicolumn{3}{|c|}{Applied approaches} & \multicolumn{2}{c|}{Sparse} & \multicolumn{2}{c|}{Dense} & {Rec.} \\ \cline{1-8}
			AT    &   SA    &          DF          & ~~2D~~ &        ~3D~        & ~2D~ &        ~3D~         &  ~3D~  \\ \hline
			&         &                      &  3.31  &        4.74        & 3.10 &        4.33         &  4.07  \\
			&         &       $\surd$        &  3.43  &        4.94        & 3.21 &        4.50         &  3.37  \\
			& $\surd$ &       $\surd$        &  3.18  &        4.51        & 3.03 &        4.17         &  3.39  \\
			$\surd$ &         &                      &  3.16  &        4.53        & 3.03 &        4.21         &  4.02  \\
			$\surd$ & $\surd$ &                      &  3.24  &        4.59        & 3.14 &        4.31         &  3.49  \\
			$\surd$ & $\surd$ &       $\surd$        &  3.05  &        4.33        & 2.93 &        4.02         &  3.25  \\ \hline
		\end{tabular}
		\vspace{-1\baselineskip}
	\end{table}

\begin{figure*}[hpt]
		\centering
		\includegraphics[width=0.96\linewidth]{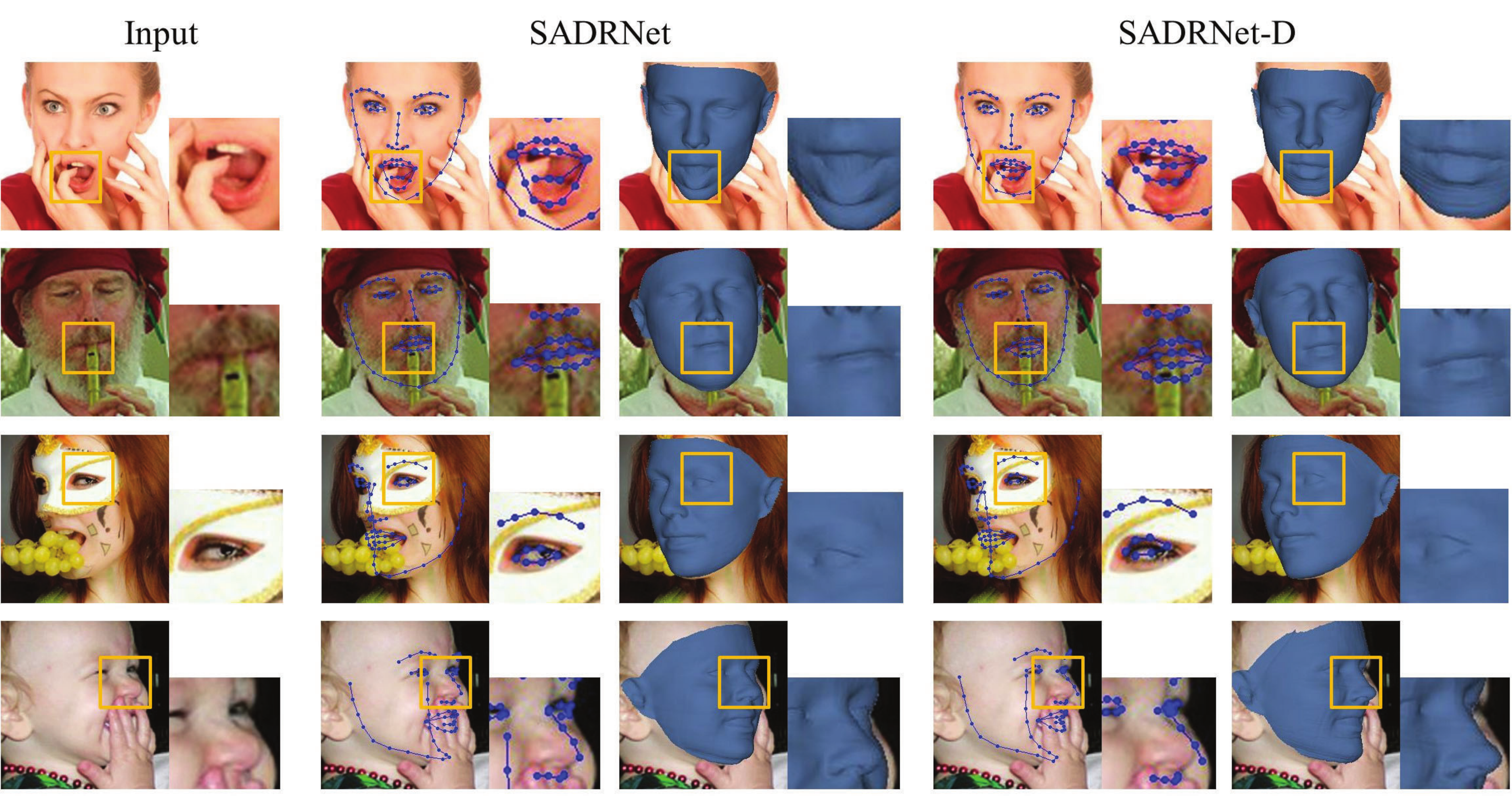}
		\caption{SADRNet vs. SADRNet-D; results for sparse alignment and face reconstruction are demonstrated. Facial regions around occlusions are zoomed in for better visual comparison.}
		\label{fig:aug}
	\end{figure*}

	{\noindent\bf Attention mechanism.}
	To study the contribution of the proposed attention mechanism, we remove the attention side branch
	from the proposed SADRNet (denoted as SADRNet-D; see Fig.~\ref{fig:ablation_framework}c for the structure.) and compare it with SADRNet. The results of the two networks are respectively summarized in the 3rd and the bottom rows in Table~\ref{tab:ablation}. We can see that the proposed attention mechanism benefits all three tasks.
	In Fig.~\ref{fig:aug}, we visualize the comparison. SADRNet-D is more easily affected by the occlusions and has lower reconstruction accuracy in the occlusion areas.
	In addition, SADRNet-D cannot estimate a face orientation as accurately as SADRNet.
	We can see the evidence from the bottom row of Fig.~\ref{fig:aug}.
	We also investigate the attention mechanism on the basic encoder-decoder architecture as shown in
	Figs.~\ref{fig:ablation_framework} a and d. The corresponding results in Table~\ref{tab:ablation} confirm
	the conclusion we draw from the comparison between SADRNet-D and SADRNet.

	{\noindent{\bf Dual face regression and self alignment.}} The distinguishing features of the self-alignment module lie in two points. One is that we obtain the target 3D face geometry through a two-stage refinement process (i.e., first pose-dependent face $\mathbf{P}$ and then the final 3D face model $\mathbf{G}$ as shown in Fig.~\ref{fig:framework}) rather than a direct regression as shown in Fig.~\ref{fig:ablation_framework}a.
	Another feature is that we estimate the pose
	information by aligning two reconstructed faces: the pose-dependent face $\mathbf{P}$ and the pose-independent face $\mathbf{S}$, rather than direct regression using the image features as in~\cite{3DDFA} (shown in Fig.~\ref{fig:ablation_framework}b).
	By comparing the first and the third result rows
	in Table~\ref{tab:ablation}, we can observe that the two-stage refinement process brings significant gains for all of the three tasks.
	The comparison between the second and third result rows demonstrates that the self-alignment-based face alignment is more reliable than that based on direct pose regression from image features. However, Fig.~\ref{fig:ablation_framework}b has a slightly better 3D face reconstruction than Fig.~\ref{fig:ablation_framework}c. This may be because
	regressing $\mathbf{S}$ and $\mathbf{P}$ together may slightly affect the estimation of $\mathbf{S}$.

	{\noindent{\bf Shape deformation.}}
	This paper regresses the pose-independent face $\mathbf{S}$ through the shape deformation, which
	estimates the differential 3D geometry relative to the mean face template. An alternative solution to regress
	$\mathbf{S}$ is to directly perform the estimation in UV space from scratch by the decoder layers as shown in Fig.~\ref{fig:ablation_framework}d.
	By comparing the results in the 5th and bottom result rows in Table~\ref{tab:ablation},
	it is easy to find that the shape regression based on deformation provides more accurate results than direct regression from scratch on all three tasks.
	Moreover, quantitatively comparing the improvement provided by using deformation (i.e., the bottom row over the 5th row) and that provided by using attention mechanism (i.e., the bottom row over the 3rd row), we can see the contribution of the deformation regression. Its improvement is more than that made by the attention mechanism in terms of 3D face reconstruction.
	
\begin{figure}[hpt]
		\centering
		\includegraphics[width=0.96\linewidth]{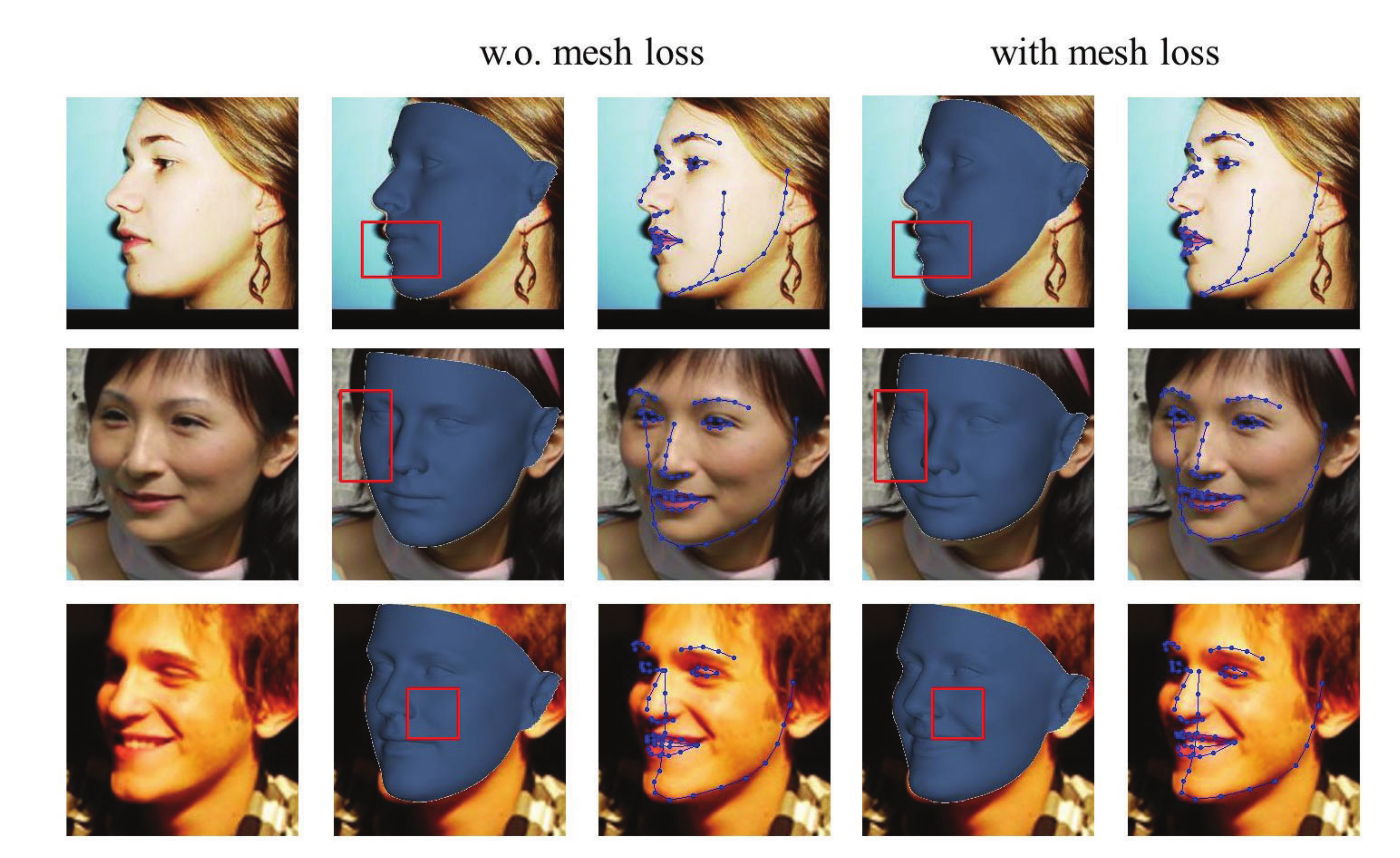}
		\caption{With mesh loss vs. without mesh loss; results for sparse alignment and face reconstruction are demonstrated.}
		\label{fig:meshloss}
	\end{figure}

	{\noindent{\bf Mesh loss.}}
	Besides the losses that directly supervise the 3D coordinates of $\mathbf{P}$, $\mathbf{S}$, and $\mathbf{G}$, inspired by~\cite{pixel2mesh}, we also adopt losses defined on the face mesh structure, i.e., the edge length loss and the normal vector loss. They do not improve the quantitative results, but help to reconstruct the face details for better visualization. In Fig.\ref{fig:meshloss}, we demonstrate some reconstruction results of the model trained with and without the mesh loss. In the demonstrated cases, the estimated landmarks are almost the same. However, the model with mesh loss can better reflect the difference between identities. In the first example with the mesh loss, the mouth's openness is more suitable. In the second example, the curvature of the cheek is better, and the reconstructed model is more recognizable. In the third example, the hollow is more obvious.
	
	\subsection{Running Time and Model Size}
	Our method has a model size of 60MB (including the mean face template parameters). In Table~\ref{tab:size} we compare the model sizes of the proposed method and the baseline methods.  Our model is lighter than all other deep-learning-based methods. The network inference for one face image takes 12ms on a GTX 1080 Ti GPU. The self-alignment post process (i.e., the step generating $\mathbf{G}$ from $\mathbf{S}$ and $\mathbf{P}$) takes 4ms on GPU or 1.5ms on an Intel Xeon E5-2690 CPU @ 2.60GHz. The fastest implementation of our method reconstructs the 3D face model from a cropped image in up to 13.5 ms.
	
	The backbone of our method, (i.e. the framework in Fig.\ref{fig:ablation_framework}.a) has a model size of 52MB. The extra size introduced by the attention side branch is 7MB. The parameters of the 7-layer pose-dependent face decoder and the 7-layer pose-independent face decoder add up to 1MB.
	
	\begin{table}
		\begin{center}
			\caption{Comparison of model size.}
			\begin{tabular}{|c|c|}
				\hline
				Method &Size\\
				\hline
				VRN~\cite{VRN} &1.5GB\\
				PRN~\cite{PRN} &153MB\\
				Nonlinear-3DMM~\cite{Nonlinear_3DMM}&152MB\\
				CMD~\cite{FPS}&93MB\\
				SADRNet (ours)&60MB\\
				
				\hline
			\end{tabular}

		\end{center}
		
		\label{tab:size}
	\end{table}
	
	\subsection{Limitations}
	We note the following limitations of our work:
	\begin{itemize}
		\item Our network only regresses the shape geometry and pose, but does not reconstruct the facial texture from the input image.
		\item The learning of the proposed SADRNet is fully supervised and depends on the costly face mesh annotation.
		Designing a weakly supervised architecture that can utilize additional data modalities (e.g., facial keypoint detection data, silhouette data, segmentation data) may improve the application potential.
		\item There is still much room for improvement in the reconstruction of high-frequency facial details (i.e.
		the pores, blemishes, and wrinkles) in our work. We believe the improvements can be made in two aspects. First, the training dataset we currently use is labeled with the parameters of 3DMM. The ground truth face models lack high-frequency details. Using datasets with high-fidelity ground truth models may help the detail synthesis. Considering that such training data is expensive, a self-supervised architecture that utilizes the facial details in the input image as supervision may be another feasible way. Second, although we have regressed the face shape deformation separately in the framework, the high-frequency details are relatively minor compared to the deformation and easy to ignore in learning. Further decomposing the shape and details or iteratively updating the face shape with cascaded regressors can more adequately supervise the learning of high-frequency details.
	\end{itemize}
	We leave these limitations as our future work.

	\section{Conclusions and Future Work}
	In this paper, we have proposed a self-aligned dual face regression network (SADRNet)  to solve the problem of 3D face reconstruction and dense alignment under unconstrained conditions. We decouple the framework of face reconstruction to two regression modules for pose-dependent and pose-independent face shape estimation, respectively. Then, a novel self-alignment module is presented to transform the detailed and more accurate face shape into its corresponding pose view to yield the final face reconstruction. To make our method robust to occlusion, we incorporate an attention module to enhance the visible facial information and estimate the transformation matrix only with visible landmarks. We evaluate our network on the AFLW2000-3D database and the Florence database. With the power of robustness to occlusions and large pose variation, our proposed method outperforms the state-of-the-art methods by a notable margin on both face alignment and 3D reconstruction.
	
	Lacking high-frequency facial details is the main drawback of our method and we consider to improve our method in the future from two aspects.
	First, we could finetune our SADRNet on a high-fidelity 3D scanned dataset or design a self-supervised learning framework to train the network with high-frequency details in the input images. Second, we plan to present a cascade reconstruction pipeline to regress our face shape in a coarse-to-fine manner and focus on more detailed face shape regression in the latter stages.

	\appendices

	\bibliographystyle{IEEEtran}
	\bibliography{new_bib2}

\end{document}